\pdfoutput=1

\documentclass{article}

\usepackage[final]{corl_2018}

\usepackage[utf8]{inputenc} % allow utf-8 input
\usepackage[T1]{fontenc}    % use 8-bit T1 fonts
\usepackage{hyperref}
\usepackage{url}            % simple URL typesetting
\usepackage{booktabs}       % professional-quality tables
\usepackage{amsfonts}       % blackboard math symbols
\usepackage{nicefrac}       % compact symbols for 1/2, etc.
\usepackage{microtype}      % microtypography
\usepackage{enumerate}
\usepackage{graphicx}
\usepackage{wrapfig}
\usepackage{paralist}
\usepackage{multirow}
\usepackage{amsmath,amsfonts,amssymb}
\usepackage{verbatim}

\newcommand{\ie}{i.e.,}
\newcommand{\etc}{etc}
\newcommand{\eg}{e.g.,}

\newcommand\mb[1]{\mathbf{#1}}
\newcommand\tb[1]{\textbf{#1}}

\graphicspath{{images/}}
\DeclareGraphicsExtensions{.png,.jpg,.pdf}

\title{Guided Feature Transformation (GFT): A Neural Language Grounding
  Module for Embodied Agents}

\author{
  Haonan Yu$^{\dagger}$, Xiaochen Lian$^{\dagger}$, Haichao
  Zhang$^{\dagger}$, and Wei Xu$^{\ddagger}$\\
  $^{\dagger}$Baidu Research, Sunnyvale CA USA\\
  $^{\ddagger}$Horizon Robotics, Cupertino CA USA\\
  \texttt{\{haonanyu,lianxiaochen,zhanghaichao\}@baidu.com,wei.xu@horizon.ai}\\
}

\begin{document}

\maketitle

\begin{abstract}
  Recently there has been a rising interest in training agents,
  embodied in virtual environments, to perform language-directed tasks
  by deep reinforcement learning.
  In this paper, we propose a simple but effective neural
  language grounding module for embodied agents that can be trained
  end to end from scratch taking raw pixels, unstructured
  linguistic commands, and sparse rewards as the inputs.
  We model the language grounding process as a
  language-guided transformation of visual features, where latent
  sentence embeddings are used as the transformation matrices.
  In several language-directed navigation tasks that feature
  challenging partial observability and require simple reasoning, our
  module significantly outperforms the state of the art.
  We also release \textsc{xworld3D}, an easy-to-customize 3D
  environment that can potentially be modified to evaluate a variety
  of embodied agents.
\end{abstract}


\section{Introduction}
This paper examines the idea of building embodied
\citep{Smith2005,Kiela2016} agents that learn control from linguistic
commands and visual inputs.
One recent line of work
\citep{Oh2017,Hermann17,Chaplot18,Das2018,Yu2018} trains such agents
situated in simulated environments in an end-to-end fashion, receiving
unstructured linguistic commands and raw image pixels as the inputs,
and producing navigation actions as the outputs.
For successful navigation control, it is crucial for an agent to learn
to associate linguistic concepts with visual features, a process known
as language \emph{grounding} \citep{Harnad1990,Siskind1994}.
To avoid a tremendous amount of labeled data, this line of work trains
language grounding oriented by navigation goals via reinforcement
learning (RL).
Through trials and errors, an agent learns not only to navigate but
also to reinforce (or weaken) the connection between visual features
and their matched (or unmatched) language tokens.

The problem of learning to control alone is quite challenging,
especially in an environment with long time horizons and sparse
rewards \citep{Sutton98}.
However, in this paper we only concentrate on language grounding as
the core of the perception system of an agent, while using an existing
RL system design, synchronous advantage actor-critic (ParallelA2C,
\citep{Clemente2017}).
Because our model exploits no structural biases of specific tasks, it
is possible to plug our language grounding module in many other neural
architectures for tasks that combine language and vision.

We propose a simple but effective neural language grounding module that
models a rich set of language-vision interactions.
The module performs a \tb{g}uided \tb{f}eature \tb{t}ransformation
(GFT), where latent sentence embeddings computed from the language
input are treated as the transformation matrices of visual features.
This guided transformation is much more expressive than the
existing three categories of language grounding methods for embodied
agents: vector concatenation \citep{Hermann17,Das2018,Shu2018}, gated
networks \citep{Chaplot18,Wu2018}, and convolutional interaction
\citep{Oh2017,Yu2018}.
In fact, it can be treated as a generalization of the last two
categories.

\begin{figure}[t]
  \resizebox{\textwidth}{!}{
    \begin{tabular}{@{}c|c@{}}
      \textbf{2D session} & \textbf{3D session}\\
      \begin{tabular}{@{}c@{\hskip 0.03in}c@{\hskip 0.03in}c@{}}
        \includegraphics[width=0.16\textwidth]{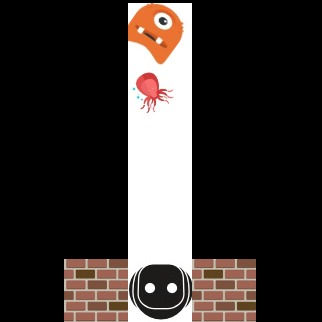}
        & \includegraphics[width=0.16\textwidth]{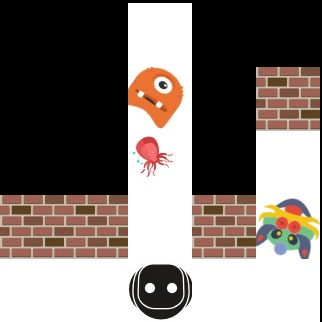}
        & \includegraphics[width=0.16\textwidth]{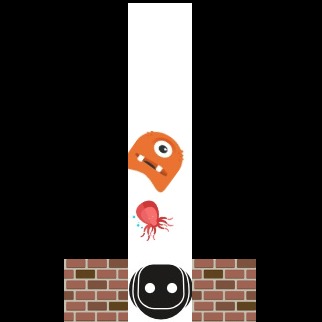}\\
      \end{tabular}
      &
      \begin{tabular}{@{}c@{\hskip 0.03in}c@{\hskip 0.03in}c@{}}
        \includegraphics[width=0.16\textwidth]{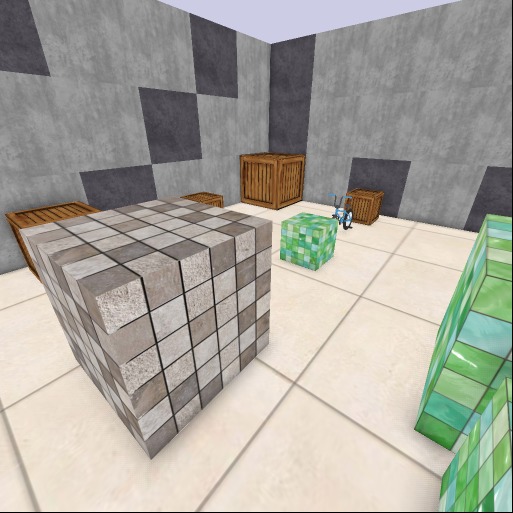}
        & \includegraphics[width=0.16\textwidth]{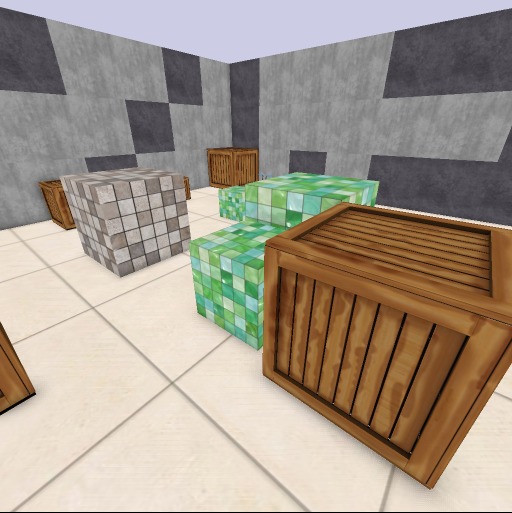}
        & \includegraphics[width=0.16\textwidth]{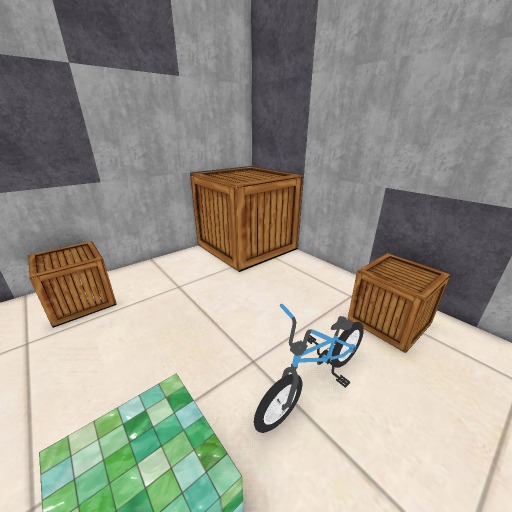}\\
      \end{tabular}\\\\
      \textit{Example command:} & \textit{Example command:}\\
      ``Navigate to the object in front of the monster.''
      & ``Can you please go to the bike?''\\
    \end{tabular}
  }
  \caption{Illustration of the two environments for evaluating
    our agent.
    The scene is randomly initialized, and only several
    key frames are shown in either case.
    The agent perceives images in the first-person view.
    The commands require language understanding skills
    such as object recognition, spatial reasoning, and semantic
    opposition.
    In the 2D case, the agent has a limited visible field regardless of
    the map size.
    The black area behind wall blocks indicates invisible regions.
    The 3D scenario with perspective distortion is closer to human
    experience.
  }
  \label{fig:envs}
\end{figure}

GFT is fully differentiable and is embedded in the perception system
of our navigation agent that is trained end to end from scratch by RL.
In an apples-to-apples comparison, our model significantly outperforms
the existing state of the art, over a rich set of navigation tasks
that feature challenging partial observability and cluttered background,
and require simple reasoning (Figure~\ref{fig:envs}).
Our GFT-powered agent is able to handle both the 2D and 3D environments
without any architecture or hyperparameter change between the two
scenarios\footnote{Video demo at \url{https://www.youtube.com/watch?v=bOBb1uhuJxg}}.
This demonstrates the generality and efficacy of GFT as a language
grounding module that can potentially benefit a variety of embodied
agents for other language-vision tasks.
Finally, we will release the \textsc{xworld3D} environment used in the
experiments.
\textsc{xworld3D} highlights a teacher infrastructure that enables
flexible customization of linguistic commands, environment maps, and
training curricula.

\section{Guided Feature Transformation for Language Grounding}
\label{sec:gfat}
The major contribution of this paper is a simple yet effective
language grounding module called GFT.
We will first describe GFT and then discuss our motivations and its
advantages.
In a general scenario, given a pair of an image $\mb{o}$ and a
sentence $\mb{l}$, a language grounding module fuses the two
modalities for downstream processings.
This is a common process seen in visual question answering (VQA)
\citep{Antol2015}, and it also lies at the heart of our agent's
perception system.

\subsection{Method}
We use a convolutional neural network (CNN) to convert $\mb{o}$ into a
feature cube $\mb{C}\in\mathbb{R}^{D\times N}$, where~$D$ is the
number of channels and $N$ is the number of locations in the image
spatial domain (collapsed from 2D to 1D for notational simplicity).
Suppose that we have an embedding function that converts $\mb{l}$ to a
series of $J$ embedding vectors
$\mb{t}_1,\ldots,\mb{t}_j,\ldots,\mb{t}_J$, where each
$\mb{t}_j\in\mathbb{R}^{D(D+1)}$ can be reshaped and treated as a
matrix $\mb{T}_j\in\mathbb{R}^{D \times (D+1)}$.
Then we compute a series of $J$ transformations, one followed by
another, activated by a nonlinear function $g$:
\begin{equation}
  \label{eq:staft}
  \mb{C}^{[j]}=g(\mb{T}_j
  \left[
  \begin{array}{l}
    \mb{C}^{[j-1]}\\
    \mb{1}^{\intercal}
  \end{array}
  \right]), \ \ \ 1 \le j \le J,
\end{equation}
where $\mb{1}\in \mathbb{R}^{N}$ is an all-one vector and
$\mb{C}^{[0]}=\mb{C}$.
This guided transformation yields a feature cube
$\mb{C}^{*}=\mb{C}^{[J]}\in\mathbb{R}^{D\times N}$ which is the final
grounding result for downstream processings.
Overall, it would be expected that the transformation matrices
$\mathbf{T}_j$ correctly capture the critical aspects of command
semantics, in order for the agent to perform tasks.
(Examples of the trained $\mathbf{T}_j$ in a later experiment are
visualized in Appendix~\ref{app:visual}.)
Despite its simple form, GFT is able to model a rich set of
interactions between language and vision, resulting in strong
representational power.
Indeed, it can be seen as a unifying formulation of two existing
language grounding modules, namely, gated networks
\citep{Perez2018,Chaplot18} and convolutional interaction
\citep{Oh2017,Yu2018}.

\subsection{Why GFT?}
\tb{GFT is a stack of \emph{generalized} gated networks.} In the
following we will ignore the subscript $j$ of $\mb{T}_j$ for
notational simplicity.
Let us first write:
\[
\mb{T}=
\left[\mb{T}'\hspace{6pt}\mb{b}\right],
\]
where $\mb{T}'\in\mathbb{R}^{D\times D}$, and $\mb{b}\in\mathbb{R}^D$
is a bias vector for the transformation in Eq.~\ref{eq:staft}.
We would like to investigate what $\mb{T}'\mb{C}$ essentially does.
Toward this end, we perform a singular value decomposition (SVD) of
$\mb{T}'$:
\[\mb{T}'=\mb{U}\mb{\Lambda}\mb{V}^{\intercal},\]
where $\mb{U}\in\mathbb{R}^{D\times D}$ and
$\mb{V}\in\mathbb{R}^{D\times D}$ are both orthogonal matrices.
$\mb{\Lambda}\in\mathbb{R}^{D\times D}$ is a diagonal matrix that
contains $D$ values $\lambda_1,\ldots,\lambda_D$ (not necessarily
non-negative) on the diagonal.

In an extreme case where both $\mb{U}$ and $\mb{V}$ are identity
matrices, each transformation step of GFT degrades to a gated
network.
Specifically, let $\mb{c}_d$ be the $d$-th feature map, we have
\begin{equation}
  \label{eq:film}
  \mb{c}_d^{[j]}=g(\lambda_d \mb{c}_d^{[j-1]} + b_d)
\end{equation}
This is exactly the same modulation provided by FiLM \citep{Perez2018}.
Intuitively, Eq.~\ref{eq:film} performs scaling, thresholding, or
negating of the features on the $d$th map of $\mb{C}^{[j-1]}$,
according to the semantics of the input command $\mb{l}$.
A further specialized version was proposed by \citet{Chaplot18} in
which they remove $g$ and $b_d$, while activating $\lambda_d$ by
sigmoid.
Thus when $\mb{U}$ and $\mb{V}$ are identities, GFT is a stack of
FiLMs.

In a general case, $\mb{U}$ and $\mb{V}$ are dense and represent
general rotations in $\mathbb{R}^D$.
Because $\mb{U}$ and $\mb{V}$ are computed from the command $\mb{l}$,
they are \emph{language-guided}.
This is a major difference between a transformation step of GFT and
that of a gated network.
The latter always modulates features in the same original feature
space, regardless of the command $\mb{l}$.
As a result, a gated network such as FiLM places more pressure on
learning $\mb{C}$ because a single feature space has to reconcile with
a huge number of commands.
When the vision is challenging or the language space is huge,
modulating only in the original feature space might become a
performance bottleneck of the overall agent model.
In contrast, a transformation step of GFT can choose to rotate the
axes of the feature space ($\mb{V}^{\intercal}$), scale the features
in that rotated space ($\mb{\Lambda}\mb{V}^{\intercal}$), and rotate
the scaled feature again ($\mb{U}\mb{\Lambda}\mb{V}^{\intercal}$).
These choices are determined by the current command.
On top of it, GFT performs this ``rotate-scale-rotate'' operation
multiple times in a sequence, when combined together, resulting in
high-order nonlinear feature modulation.

\tb{GFT performs concept detection over \emph{multiple} convolutional
  steps.} An alternative interpretation is possible if we treat
Eq.~\ref{eq:staft} as a $1 \times 1$ convolution with $D$ filters and
a stride of one, sharing similar motivations in \citet{Oh2017} and
\citet{Yu2018}.
In such an interpretation, each row of $\mb{T}$ contains a $1\times
1$ convolutional filter of length $D$ and a scalar bias.
The $D$ filters (rows) represent at most $D$ different or
complementary aspects of the semantics of the command $\mb{l}$, and
a step of Eq.~\ref{eq:staft} essentially performs concept detection.
For example, a 3D asymmetric object such as a bike has different
appearances from different viewing angles.
Thus having multiple filters of the sentence ``go to bike'' improves
the representational power and results in a higher chance of finding
the corresponding concept when the agent moves around.
Overall, GFT performs multi-step concept detection with
language-dependent filters at each step.

In summary, the simple but general formulation of GFT unifies several
existing ideas for grounding language in vision.
In the remainder of this paper, we will evaluate it in a challenging
language-directed navigation problem.

\section{The Navigation Problem}
\label{sec:problem}
We formally introduce our problem as a partially observable Markov
decision process (POMDP) \citep{Kaelbling1998} as follows.
The problem is divided into many navigation sessions.
At the beginning of each session, a navigation task is sampled by
a pre-programmed teacher as $k\sim P(k)$.
Given the task, an initial environment state $\mb{s}^{[1]}$ is sampled
by the simulator as $\mb{s}^{[1]}\sim P(\mb{s}^{[1]}|k)$, \ie\ the
simulator arranges the scene according to the sampled task.
An environment state $\mb{s}$ contains both the map configuration and
the agent's pose.
We assume that the teacher has full access to the environment state,
and samples a linguistic command $\mb{l}\sim P(\mb{l}|\mb{s}^{[1]},k)$
which will be used throughout the session.
At each time step $t$, the agent only has a partial observation of the
environment, computed by a rendering function
$\mb{o}^{[t]}=o(\mb{s}^{[t]})$.
We assume that $P(k)$, $P(\mb{s}^{[1]}|k)$,
$P(\mb{l}|\mb{s}^{[1]},k)$, and $o(\cdot)$ are all unknown to the agent.

Below we use $\theta$ to denote the complete set of the model
parameters, and let any function that uses a subset of parameters
directly depend on $\theta$ for notational simplicity.
The agent takes an action $a^{[t]}$ according to its policy
$\pi_{\theta}$, given the current observation $\mb{o}^{[t]}$, the
history $\mb{h}^{[t-1]}$, and the command $\mb{l}$:
\begin{equation}
  a^{[t]} \sim \pi_{\theta}(a^{[t]}|\mb{o}^{[t]}, \mb{h}^{[t-1]}, \mb{l}),
\end{equation}
where a latent vector
$\mb{h}^{[t-1]}=h_{\theta}(\mb{o}^{[1:t-1]},a^{[1:t-1]},\mb{l})$
summarizes all the previous history of the agent at time $t$ in the
current session.
The teacher observes the action and gives the agent a scalar reward
$r^{[t]}=r(a^{[t]},\mb{s}^{[t]},\mb{l})$.
Note that the reward depends not only on the state and the action, but
also on the command.
The environment then transitions to a new state $\mb{s}^{[t+1]}\sim
P(\mb{s}^{[t+1]}|\mb{s}^{[t]},a^{[t]})$.
This act-and-transition iteration goes on until a terminal state or
the maximum time $T$ is reached.
Our problem is to maximize the expected reward:
\begin{equation}
  \label{eq:obj}
  \max\limits_{\theta}\mathbb{E}_{\mathcal{S},\mathcal{A},\mb{l}}\left[\sum\limits_t\gamma^{t-1}r^{[t]}\right],
\end{equation}
where $\mathcal{S}=(\mb{s}^{[1]},\ldots,\mb{s}^{[t]},\ldots)$ and
$\mathcal{A}=(a^{[1]},\ldots,a^{[t]},\ldots)$ denote the state and
action sequences, respectively.
$\gamma\in[0,1]$ is a discount factor.
Note that $r(\cdot)$ and $P(\mb{s}^{[t+1]}|\mb{s}^{[t]},a^{[t]})$ are
also unknown to the agent, namely, the RL agent is model-free.

\begin{figure}[t]
  \begin{center}
    \resizebox{0.95\textwidth}{!}{
      \includegraphics[width=\textwidth]{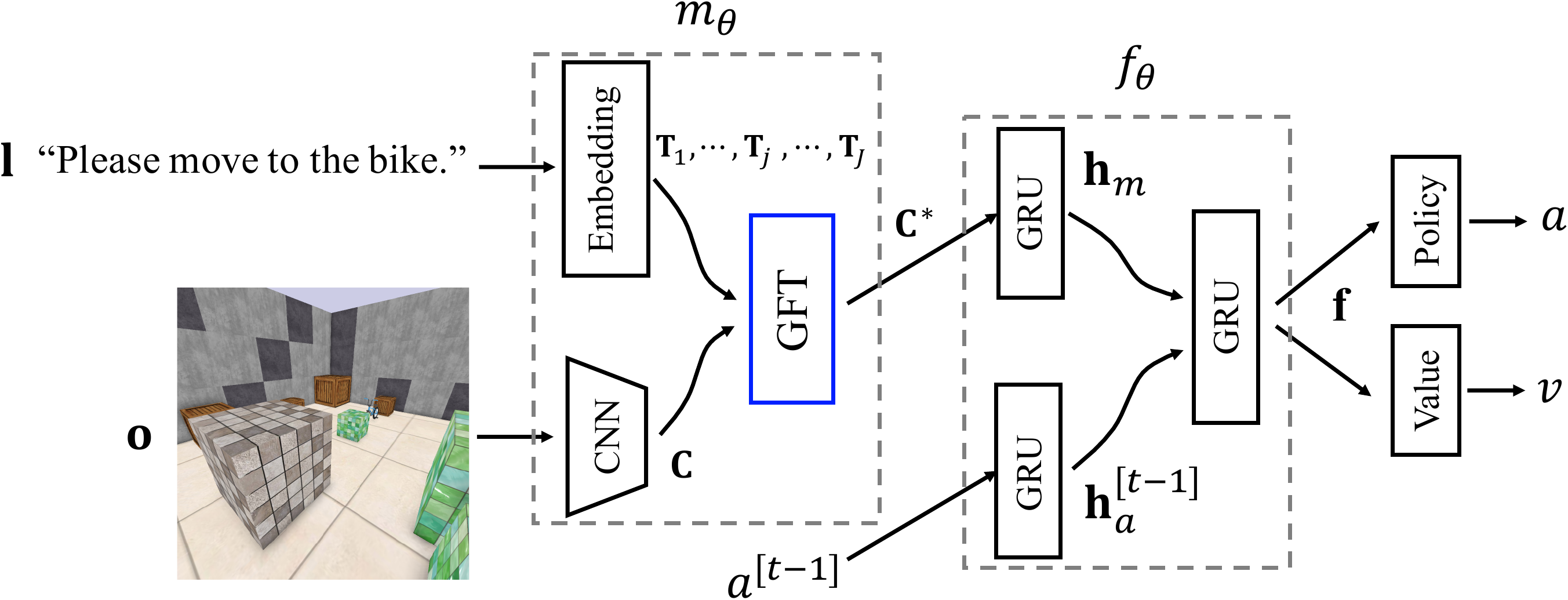}
    }
    \caption{An overview of our agent architecture.
      In general, it is a two-layer GRU network with GFT embedded
      as the core for perception, and with value and policy networks
      for control.
      For notational simplicity, we only keep the superscript $t-1$
      to indicate quantities from the previous time step, while
      removing the superscript $t$ for quantities at the current
      time step.
      The two GRU states $\mb{h}_m^{[t-1]}$ and $\mb{h}_a^{[t-1]}$
      (defined in Eq.~\ref{eq:arch} Appendix~\ref{app:ga2c}) together
      form the agent history $\mb{h}^{[t-1]}$.
      From the figure we can see that GFT is a portable module that
      could potentially be plugged into other networks involving
      language-vision interactions.
    }
    \label{fig:overview}
    \vspace{-2ex}
  \end{center}
\end{figure}

The objective Eq.~\ref{eq:obj} fits in the standard RL framework and
is readily solvable by the actor-critic (AC)
algorithm~\citep{Sutton1999}.
Specifically, we compute the following policy gradient for any time step $t$:
\begin{equation}
  \label{eq:gradient}
  -\mathbb{E}_{\mb{s}^{[t]},a^{[t]},\mb{l}}
  \left[\left(\nabla_{\theta}\log\pi_{\theta}(a^{[t]}|\mb{o}^{[t]},\mb{h}^{[t-1]},\mb{l})
    +\eta\nabla_{\theta}v_{\theta}(\mb{o}^{[t]},\mb{h}^{[t-1]},\mb{l})\right)A^{[t]}
    + \kappa\nabla_{\theta}\mathcal{E}(\pi_{\theta})\right],
\end{equation}
where $v_{\theta}$ is the estimated value function,
$\mathcal{E}(\cdot)$ denotes entropy for encouraging exploration
\citep{Mnih2016}, and $\kappa>0$ and $\eta>0$ are constant weights.
The advantage $A^{[t]}$ is computed as
\begin{equation}
  \label{eq:advantage}
  A^{[t]}=r^{[t]}+\gamma
  v_{\theta}(\mb{o}^{[t+1]},\mb{h}^{[t]},\mb{l})-v_{\theta}(\mb{o}^{[t]},\mb{h}^{[t-1]},\mb{l}).
\end{equation}
Our implementation adopts the ParallelA2C design \citep{Clemente2017}
to aggregate a minibatch of gradients (Eq.~\ref{eq:gradient}) over
multiple identical agents running in parallel (each agent in a
separate copy of the environment) over multiple time steps, with
$\theta$ synchronized and shared among their models.

Note in Eq.~\ref{eq:gradient} that the inputs of the policy network
$\pi_{\theta}$ and the value network $v_{\theta}$ are identical.
Thus we share a sub-network between $\pi_{\theta}$ and $v_{\theta}$
for parameter efficiency \citep{Mnih2016}.
The sub-network outputs a latent state representation $\mb{f}^{[t]}$
and has two stages:
\begin{equation}
  \mb{f}^{[t]}
  =f_{\theta}\left(m_{\theta}(\mb{o}^{[t]},\mb{l}),\mb{h}^{[t-1]}\right).
\end{equation}
The first stage $m_{\theta}$ is a multimodal function that grounds
language in vision, and the second stage $f_{\theta}$ combines the
grounding result with the agent history $\mb{h}^{[t-1]}$.
We instantiate $m_{\theta}$ by using GFT for language grounding, and
instantiate $f_{\theta}$ as a gated recurrent unit (GRU)
\citep{Cho2014}.
An overview of the agent architecture is illustrated in
Figure~\ref{fig:overview}.
We refer the reader to more details in Appendix~\ref{app:ga2c}.

\section{Related Work}
\tb{Virtual navigation.} Prior to this work, there have been
several studies demonstrating virtual agents learning to navigate in
virtual environments, based on reinforcement signals
\citep{Jaderberg2017,Mirowski2017}.
Despite the impressive results achieved, these studies usually have
fixed goals for agents.
For example, an agent always learns to pick up apples or avoid
enemies with specific appearances.
In other words, the agent's goals are fixed and cannot be changed,
unless the rewards are modified followed by retraining.
There is no language understanding involved: the perceptual inputs are
images only.

\tb{Multi-goal virtual navigation.} A recent line of work augments the
above virtual navigation with multiple non-linguistic goals.
These goals are specified in different forms: target images
\citep{Zhu2017a,Zhu2017b}, one-hot or continuous embeddings
\citep{Brahmbhatt2017,Gupta2017a,Oh2017,Savva2017,Mirowski2018},
target poses \citep{Gupta2017b}, etc.
In contrast, our focus is on understanding linguistic commands.

\tb{Language-directed virtual navigation.} Another recent line of work
\citep{Hermann17,Oh2017,Chaplot18,Das2018,Yu2018,Anderson2018,Gordon2018}
augments the virtual navigation problem with linguistic inputs, where
an agent's goal always depends on an instructed command.
Accordingly, it is crucial for these methods to ground language in vision.
Our GFT generalizes and improves some existing language grounding
modules (details in Section~\ref{sec:gfat}), while incurring negligible
additional costs in implementation and training time.

\tb{Visual question answering.} Unlike VQA
\citep{Antol2015,Lu2016,Yang2016,Perez2018,Das2018,Gordon2018}, our
problem does not require the agent to answer questions.
Instead, the agent takes movement actions to respond to the teacher.
However, both problems require language grounding, the study of which
might be transferred from one problem to another.
Indeed, in Section~\ref{sec:exp}, the \tb{FiLM} comparison module is
adapted from \citet{Perez2018}, the \tb{CGated} and \tb{Concat}
modules were adopted in some early work on VQA \citep{Antol2015}, and
the \tb{Concept} comparison module resembles the stacked attention
network (SAN) \citep{Yang2016} when there is only one single attention
layer.
Although GFT is proposed for embodied agents, we hope that it will
also benefit research on VQA.

\tb{Grounding language in vision and robotics.} Our work is also
related to language grounding in realistic
images~\citep{Yu2013,Gao2016,Rohrbach2016} and robotics navigation
under language \citep{Chen2011,Tellex2011,Barrett2017}, where static
labeled datasets are required.
In addition, these methods for language understanding usually employ
structural assumptions specifically for their problems.
In contrast, our GFT module is general-purpose and could potentially
be easily applied to a wide range of problems that require language
grounding.

\section{Experiments}
\label{sec:exp}
We evaluate our agent in two challenging environments:
\textsc{xworld2D} and \textsc{xworld3D} (Figure~\ref{fig:envs}).
Both environments host the same set of five types of language-directed
navigation tasks described in Table~\ref{tab:tasks}.
A common syntax is shared by the two environments for generating task
commands.
Except object words, the remaining lexicon including grammatical and
spatial-relation words, is also shared.
Thus the only differences between \textsc{xworld2D} and
\textsc{xworld3D} in our experiments are graphics and objects.
Both environments generate random navigation sessions following the
problem definition in Section~\ref{sec:problem}.

Despite the huge difference between the visual structures of the two
worlds, we apply a structurally identical agent to both of them.
This identity includes the same network architecture, the same set of
actions (\texttt{move\_forward}, \texttt{move\_backward},
\texttt{move\_left}, \texttt{move\_right}, \texttt{turn\_left}, and
\texttt{turn\_right}), and the same set of hyperparameter values
(\eg\ learning rate, batch size, momentum, layer sizes, \etc).
Only the model parameters are different and learned separately.
Such an experiment setting tests the generalizability, efficacy, and
portability of GFT as a language grounding module.

\begin{table}[t!]
  \centering
  \resizebox{\textwidth}{!}{
    \begin{tabular}{l|l|l}
      \textbf{Type} & \textbf{Navigation target} & \textbf{Example command}\\
      \hline
      \texttt{nav} & the specified object & ``Please go to the chair.''\\
      \texttt{nav\_nr} & an object near the specified one &
      ``Move to the object near the chair.''\\
      \texttt{nav\_bw} & the location between the two objects &
      ``Go to the location between the chair and the table?''\\
      \texttt{nav\_avoid} & any object but the specified one &
      ``Avoid the chair.''\\
      \texttt{nav\_dir} & an object specified by a
      relative & ``Navigate to the object left of the chair.''\\
      & direction w.r.t. another object &\\
    \end{tabular}
  }
  \caption{The five types of navigation tasks in both environments.}
  \vspace{-4ex}
  \label{tab:tasks}
\end{table}

\subsection{Comparison Methods}
We perform an apples-to-apples comparison with six state-of-the-art
language grounding modules for embodied agents.
To do so, we make minimal changes to our agent architecture when
implementing the comparison methods:
only the multimodal function $m_{\theta}$ is changed for each method,
with the remaining components unchanged.
Regardless of the choice, the output of $m_{\theta}$ is always
flattened to a vector as an input to $f_{\theta}$.
We assume that the input command is always first encoded into a
fixed-length embedding $\mb{l}_{\text{BoW}}$ by a bag-of-words (BoW)
encoder for training efficiency.
The six comparison methods are:
\begin{compactenum}[]
\item[\tb{Concat}] \citep{Hermann17,Das2018,Shu2018,Gordon2018}
  directly concatenates a compact language embedding and a compact
  visual embedding.
  We project~\footnote{In the remainder of this paper, a projection
  denotes a fully-connected (FC) layer followed by a nonlinear
  activation function.}  both $\mb{l}_{\text{BoW}}$ and $\mb{C}$ to
    the same dimension before the concatenation.
\item[\tb{Gated}] \citep{Chaplot18} weights the feature maps of
  $\mb{C}\in \mathbb{R}^{D\times N}$ by a gate vector
  $\mb{l}_{\text{gate}}\in [0,1]^{D}$.
  In our case, $\mb{l}_{\text{gate}}$ is generated by a two-layer
  multilayer perceptron (MLP) from $\mb{l}_{\text{BoW}}$.
\item[\tb{CGated}] \citep{Wu2018} is a variant of \tb{Gated}.
  Instead of weighting feature maps of $\mb{C}$, they project
  $\mb{C}$ down to a visual embedding which is then weighted by
  $\mb{l}_{\text{gate}}$, a gate vector projected from
  $\mb{l}_{\text{BoW}}$ to the same dimension of the visual embedding.
\item[\tb{FiLM}] \citep{Perez2018} follows exactly Eq.~\ref{eq:film}
  which can be seen as a special case of our method.
  All $\lambda_{d}$ and $b_{d}$ are generated by a two-layer MLP
  from $\mb{l}_{\text{BoW}}$.
\item[\tb{Concept}] \citep{Yu2018} directly treats $\mb{l}_{\text{BoW}}$
  as a $1 \times 1$ filter.
  An attention map is obtained by convolving $\mb{C}$ with the filter.
  In addition, $\mb{C}$ is convolved with a $1\times 1$ filter at a
  stride of one to produce an environment map.
  Finally, the attention map and the environment map are concatenated.
\item[\tb{EncDec}] \citep{Anderson2018} makes several modifications
  to the original implementation to better suit our problem.
  First, we train the CNN from scratch.
  Second, we compute the instruction context directly as
  $\mb{l}_{\text{BoW}}$ without using word attention, since our
  teacher will not issue detailed, long-paragraph commands.
  Third, we add one additional layer of GRU after the concatenation
  of the decoder state and the instruction context, to model longer-range
  temporal dependency.
\end{compactenum}
Two variants of our agent are reported: \tb{GFT-1} ($J=1$) and
\tb{GFT-2} ($J=2$)\footnote{According to our observation, $J=3$ appears
  to be a saturation point whose performance has almost no gain over
$J=2$.}.
We generate $\mb{T}_j$ by a two-layer MLP from $\mb{l}_{\text{BoW}}$.
After training, each of the eight methods is evaluated for 10k test
sessions, where the models saved for the final three passes (each pass
contains 5k minibatches) of each method are used to obtain an average
result.
The comparison setting described in this section applies to both
\textsc{xworld2D} and \textsc{xworld3D}.
More details of the comparison methods and our method are described in
Appendix~\ref{app:model}.

\subsection{Optimization Details}
The optimization details described in this section apply to all the
methods in both environments.
We adopt RMSprop \citep{Tieleman2012} with a learning rate of
$10^{-5}$, a damping factor of $\epsilon=0.01$, and a gradient moving
average decay of $\rho=0.95$.
The gradient has a momentum of 0.9.
The batch size is set to $128$.
The total number of training batches is 2 million.
The parameters of each method are initialized with four different
random seeds, the results of which are averaged and reported.

\subsection{Results for \textsc{xworld2D}}
\tb{Environment and action.} We modified the \textsc{xworld2D}
environment \citep{Yu2018} to host our navigation tasks.
The original fully-observable setting now becomes a
partially-observable egocentric setting in 2D.
The original action set is augmented with \texttt{turn\_left} and
\texttt{turn\_right}, of which the yaw changes are both $90^{\circ}$.
To increase the visual variance, at each session we randomly
rotate each object and scale it randomly within $[0.5,1.0]$.
Suppose each map is $X\times Y$, then a session will end after $3XY$
time steps if a success or failure is not achieved.
In the experiment, we set $X=Y=8$.
Each $8\times 8$ map contains 4 objects and 16 obstacles.
We use Prim's algorithm \citep{Prim1957} to randomly generate a
minimal spanning tree for placing the obstacles so that the map is
always a valid maze.
The objects and the agent are then randomly initialized while
complying with the sampled navigation task $k$.
The agent can only see a $5\times 5$ area in front of it, excluding
any region occluded by obstacles.
Generalization to larger maps will be investigated in
Appendix~\ref{app:generalization}.

\tb{Rewards.} A success (failure) according to the teacher's command
gives the agent a $+1$ ($-1$) reward.
A failure is triggered whenever the agent hits any object that is not
required by the command.
The time penalty of each step is $-0.01$.
No other extrinsic or intrinsic rewards are used.

\tb{Objects and vocabulary.} We use a collection of 345 object
instances released by \citet{Yu2018}, constituting 115 object
classes in total.
The vocabulary contains 115 object words, 8 spatial-relation words,
and 40 grammatical words, for a total size of 163.
In total, there are 1,187,850 distinct sentences that can be generated
by the teacher's predefined context-free grammar (CFG).
The lengths of these sentences range from 1 to 15.

\begin{figure}[t]
  \begin{center}
    \begin{tabular}{@{}cc@{}}
      \includegraphics[width=0.48\textwidth]{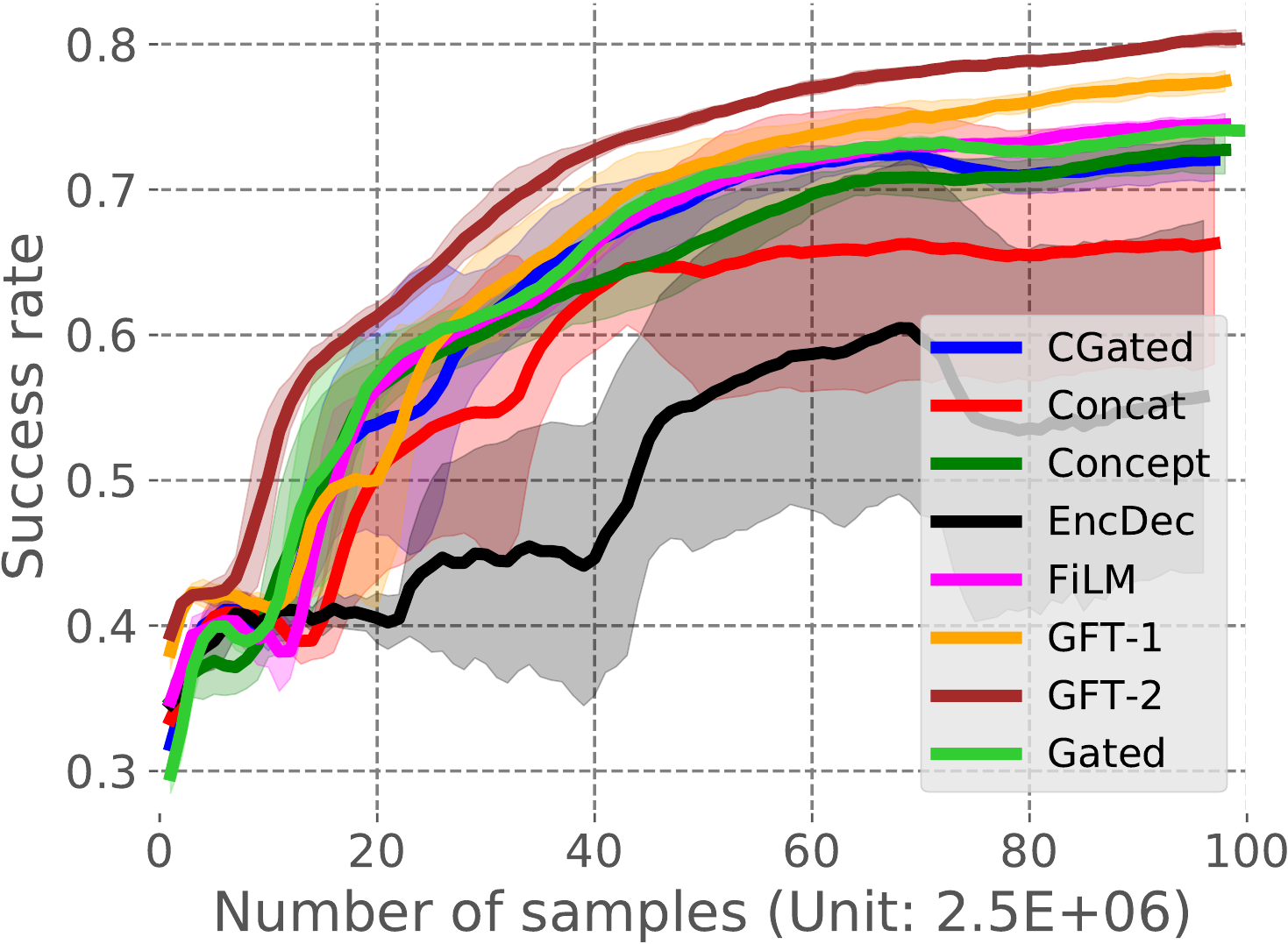}
      & \includegraphics[width=0.48\textwidth]{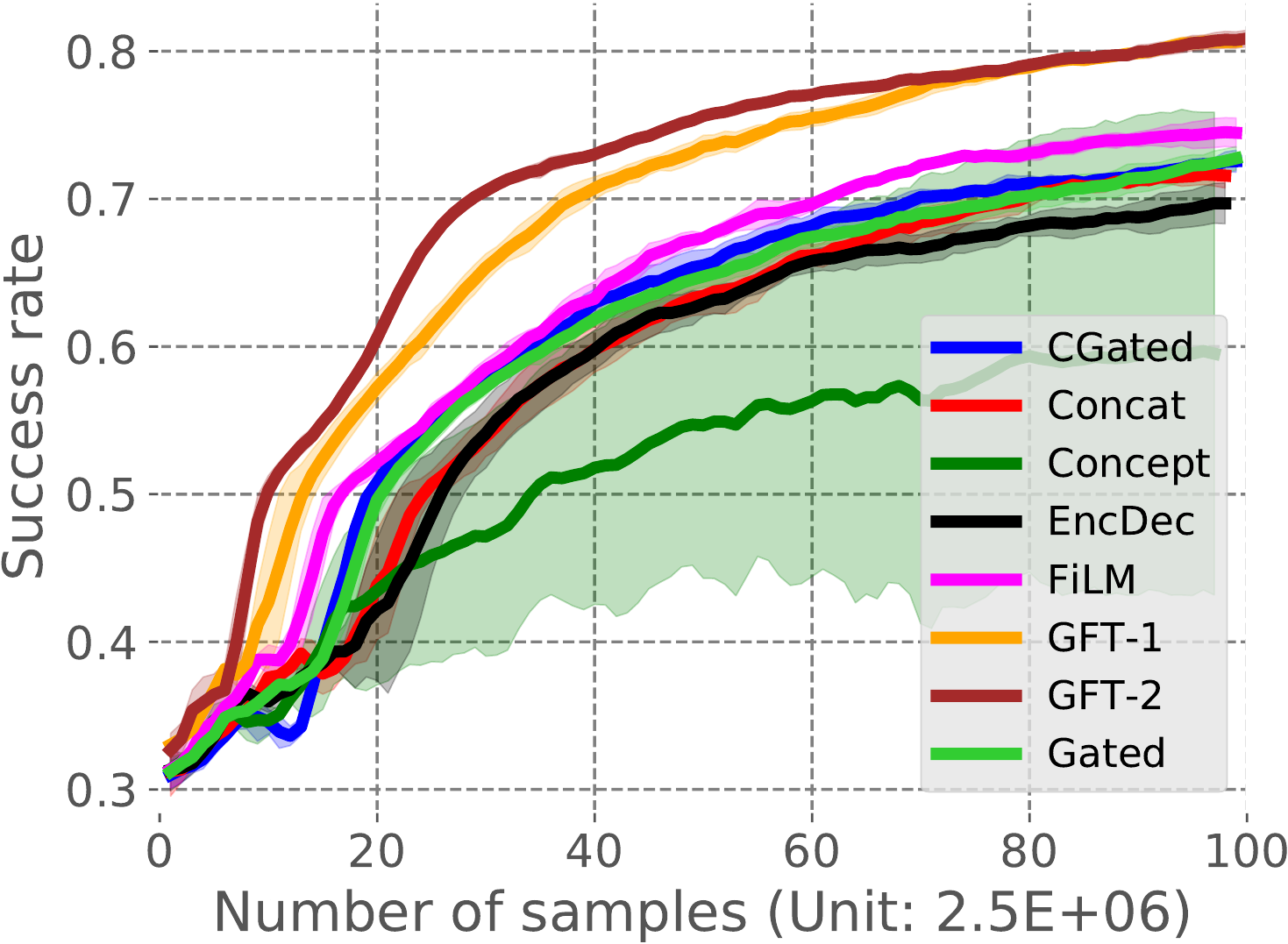}\\
      (a) \textsc{xworld2D} & (b) \textsc{xworld3D}\\
    \end{tabular}
    \caption{Success rates (averaged over four random restarts)
      vs.\ number of training samples (time steps).
      The shaded area around each curve denotes the standard deviation.
      Although each method trains the same number of minibatches, for
      details explained in Appendix~\ref{app:ga2c}, the total number
      of actions taken by the agent might vary slightly for different
      methods.
    }
    \label{fig:curves}
  \end{center}
\end{figure}

\tb{Results.} The training curves of success rates are shown in
Figure~\ref{fig:curves} (a).
We observe that \tb{GFT-1} has some marginal improvement on the
success rates of the best-performing comparison methods such as
\tb{FiLM} and \tb{Gated}.
As we perform a second feature transformation, \tb{GFT-2} produces a
performance jump.
Our explanation is that the visual recognition challenge, with random
object yaws and scales in each session, requires an expressive
language grounding function that can be better modeled by multiple
steps of Eq.~\ref{eq:staft}.
Table~\ref{tab:rates} (2D) shows the test results split into five
navigation types.
\tb{GFT-2} produces the best numbers in all the five columns.
Unsurprisingly, \texttt{nav\_avoid} has the highest rate because the
agent only has to go to an arbitrary target which is not the specified
one.
\texttt{nav} has the lowest rate because the agent cannot exploit any
object arrangement pattern like in \texttt{nav\_bw}.
See Figure~\ref{fig:2d-examples} Appendix~\ref{app:examples} for
example navigation sessions.

\subsection{Results for \textsc{xworld3D}}
\tb{Environment and action.} The environment layout and the agent's
action are both discrete in \textsc{xworld3D}.
An $X\times Y$ map consists of $XY$ square grids, each grid as a unit
containing an object, an obstacle, or nothing.
An object always has a unit scale and a random yaw when initialized.
An obstacle has one of four scales: $1.0$, $0.7$, $0.5$, and $0.3$, which
is randomly sampled when the obstacle is initialized.
The agent walks (\texttt{move\_\{forward,backward,left,right\}})
roughly half of a grid per time step.
The yaw change of the agent when it turns
(\texttt{turn\_\{left,right\}}) is $45^{\circ}$ per time step.
A session will end after $10XY$ time steps if a success or failure is
not achieved.
In the experiment, we set $X=Y=8$.
Each $8\times 8$ map contains 4 objects and 16 obstacles, and its
initialization follows the same process in the 2D case.
Generalization to larger maps will be investigated in
Appendix~\ref{app:generalization}.

\tb{Rewards.} The reward function is the same with the 2D case.

% grammatical words:
% ['.', '?', '!', 'well', 'done', 'end', 'time', 'up',
% 'please', 'is', 'your', 'the', 'go', 'to', 'navigate', 'reach',
% 'move', 'could', 'collect', 'you', 'can', 'will', 'destination',
% 'target', 'goal', 'wrong', 'object', 'of', 'in', 'that', 'do',
% 'not', 'avoid', 'anything', 'except', 'but', 'and', 'location',
% 'place', 'grid']

% object words:
% ['cat', 'dog', 'rooster', 'bed', 'sofa', 'fan',
% 'staircase', 'vase', 'barbecue', 'bench', 'cooker', 'oven', 'table',
% 'wardrobe', 'barrel', 'bookshelf', 'drawers', 'screen', 'trashcan',
% 'books', 'firehydrant', 'plant', 'umbrella', 'drums',
% 'mailbox', 'tricycle', 'wheelchair', 'bottle', 'apples', 'car',
% 'flashlight', 'photo', 'sunglasses', 'backpack', 'carpet',
% 'flowers', 'piano', 'table-tennis', 'basket', 'chair', 'fountain',
% 'pillow', 'toilet', 'basketball', 'chessboard', 'gift', 'towel',
% 'bathtub', 'clock', 'guitar', 'pool-table', 'train', 'bike',
% 'hair-dryer', 'puzzle', 'trampoline', 'boiler', 'headphones',
% 'rabbit', 'treadmill', 'bread', 'comb', 'horse', 'brush', 'crib',
% 'iron', 'scale', 'bucket', 'cup', 'scissors', 'vacuum', 'burger',
% 'dart', 'lamp', 'slippers', 'cake', 'dog-house', 'laptop',
% 'calender', 'milk', 'speaker', 'camera', 'fence', 'pan', 'squeezer',
% 'candle', 'phone', 'stove']

% spatial relations:
% ['near', 'by', 'besides', 'left', 'right', 'front', 'behind',
% 'between']

\tb{Objects and vocabulary.} \textsc{xworld3D} contains 88 different
objects\footnote{Downloaded from
  \url{http://www.sweethome3d.com/freeModels.jsp}}.
There are three types of obstacles: brick, crate, and cube.
The vocabulary is the same with the 2D case, except that there are 88
different object words and the vocabulary size is now 136.
In total, there are 709,383 distinct sentences that can be generated
by the teacher according to the same set of syntax rules in the 2D
case.
The lengths of these sentences also range from 1 to 15.

\tb{Results.} The training curves of success rates are shown in
Figure~\ref{fig:curves} (b).
We observe that \tb{GFT-1} already has a huge advantage over the
best-performing comparison method \tb{FiLM}.
On top of this, \tb{GFT-2} shows a faster performance increase
during the training time.
This suggests that the original feature space cannot easily comply
with various language commands.
A rotation of the feature space depending on the input command, an
operation which \tb{FiLM} lacks but \tb{GFT-1} owns, is important for
producing better grounding results.
Table~\ref{tab:rates} (3D) shows the test results split into five
navigation types.
Unlike in the 2D case, now for \tb{GFT-2}, \texttt{nav} has the
second-best success rate while \texttt{nav\_dir} has the lowest rate.
Visualization of the \texttt{nav\_dir} test cases reveals that due to
severe perspective distortion in 3D, the agent has some difficulty of
grounding the spatial-relation words, especially when multiple objects
are located nearby.
But still, GFT agents obtain much better \texttt{nav\_dir} results
than the comparison methods (over a 15\% increase on average).
Several example navigation sessions are shown in
Figure~\ref{fig:3d-examples} Appendix~\ref{app:examples}.

\begin{table}[t!]
  \centering
  \resizebox{\textwidth}{!}{
  \begin{tabular}{l|c|c|c|c|c|c|c|c|c|c}
    & \multicolumn{2}{c|}{\texttt{nav}}
    & \multicolumn{2}{c|}{\texttt{nav\_avoid}}
    & \multicolumn{2}{c|}{\texttt{nav\_bw}}
    & \multicolumn{2}{c|}{\texttt{nav\_dir}}
    & \multicolumn{2}{c}{\texttt{nav\_near}}
    \\
%    \cline{2-11}
    & 2D & 3D
    & 2D & 3D
    & 2D & 3D
    & 2D & 3D
    & 2D & 3D
    \\
    \hline
    \tb{Concat}
    & 54.1$\pm$3.7
    & 73.4$\pm$3.4
    & 90.7$\pm$2.8
    & 94.0$\pm$2.2
    & 77.8$\pm$5.4
    & 78.6$\pm$3.2
    & 53.1$\pm$30.1
    & 47.6$\pm$3.9
    & 58.5$\pm$11.8
    & 62.1$\pm$2.9
    \\
    \tb{Gated}
    & 58.1$\pm$3.8
    & 74.2$\pm$3.8
    & 90.6$\pm$2.3
    & 94.2$\pm$2.8
    & 84.0$\pm$3.9
    & 80.4$\pm$5.2
    & 74.5$\pm$4.1
    & 48.1$\pm$4.9
    & 63.7$\pm$3.7
    & 61.5$\pm$5.0
    \\
    \tb{CGated}
    & 58.7$\pm$3.9
    & 74.0$\pm$2.9
    & 91.2$\pm$1.9
    & 95.1$\pm$1.9
    & 84.6$\pm$3.1
    & 80.6$\pm$3.8
    & 69.3$\pm$4.2
    & 47.7$\pm$4.6
    & 63.4$\pm$3.2
    & 64.8$\pm$4.0
    \\
    \tb{FiLM}
    & 58.8$\pm$3.8
    & 78.6$\pm$2.9
    & 91.0$\pm$2.4
    & 95.4$\pm$1.8
    & 83.7$\pm$2.9
    & 78.6$\pm$4.2
    & 73.2$\pm$4.3
    & 52.0$\pm$4.6
    & 66.4$\pm$5.6
    & 68.0$\pm$3.7
    \\
    \tb{Concept}
    & 64.6$\pm$4.3
    & 65.8$\pm$19.1
    & 93.4$\pm$1.6
    & 90.5$\pm$10.8
    & 80.8$\pm$3.0
    & 60.9$\pm$24.4
    & 61.5$\pm$3.9
    & 31.7$\pm$13.7
    & 64.9$\pm$3.9
    & 54.0$\pm$16.7
    \\
    \tb{EncDec}
    & 51.5$\pm$7.4
    & 67.1$\pm$5.1
    & 84.4$\pm$11.0
    & 92.7$\pm$1.9
    & 73.6$\pm$12.1
    & 77.7$\pm$3.9
    & 20.5$\pm$27.4
    & 48.9$\pm$3.8
    & 50.9$\pm$12.2
    & 59.1$\pm$4.1
    \\
    \hline
    \tb{GFT-1}
    & 65.0$\pm$4.2
    & \tb{85.2}$\pm$3.4
    & 93.8$\pm$1.9
    & \tb{96.3}$\pm$1.3
    & 84.7$\pm$3.5
    & \tb{82.1}$\pm$5.3
    & 76.4$\pm$3.1
    & 61.2$\pm$3.9
    & 70.4$\pm$4.2
    & 78.3$\pm$4.5
    \\
    \tb{GFT-2}
    & \tb{70.7}$\pm$3.5
    & 84.6$\pm$3.0
    & \tb{94.5}$\pm$1.8
    & \tb{96.3}$\pm$1.9
    & \tb{85.6}$\pm$2.3
    & 80.5$\pm$2.7
    & \tb{80.7}$\pm$2.6
    & \tb{61.4}$\pm$4.9
    & \tb{72.3}$\pm$3.7
    & \tb{78.8}$\pm$2.6
    \\
  \end{tabular}
  }
  \caption{The evaluation results for 10k test sessions.
    The average navigation success rates are reported as percentages.
    Numbers in \tb{bold} represent the best ones.
    The second number in each cell represents the standard deviation
    over twelve test runs, each run corresponding to one pass (out of
    three final passes) of one trained model (out of four random
    initializations).
  }
  \label{tab:rates}
  \vspace{-4ex}
\end{table}

\begin{comment}
\subsection{Testing on Open-space Maps}
%
To see how obstacles affect a trained agent's navigation, for
\tb{GFT-1} and \tb{GFT-2} we perform additional tests with the
trained models, where obstacles are removed to form open space.
%
The last two rows of Table~\ref{tab:rates} show the performance on
open-space maps.
%
As expected, we observe that removing obstacles significantly improves
the success rates.
%
One exception is \texttt{nav\_dir} in \textsc{xworld2D}, where the
performance drops instead.
%
Explain ...
\end{comment}

\subsection{Limitations}
While GFT is theoretically more expressive and empirically better than
some of the existing language grounding modules, there are certain
limitations of it.
First, because each $\mb{T}_j$ is generated from the command, it
requires a large projection matrix.
In our implementation, the projection matrix that converts a hidden
sentence embedding of length 128 to $\mb{T}_j$ has a size of
$128\times D(D+1)=128\times 4160$ assuming $D=64$.
Thus usually GFT has more parameters to be learned compared with its
simplified versions like FiLM.
An alternative might be to explicitly constrain $\mb{T}_j$ to be
sparse and possibly low-rank.
Second, GFT performs several steps of transformations, each of which
has a different transformation matrix.
This further linearly increases the number of learnable parameters.
One solution would be to set
$\mb{T}_1=\ldots=\mb{T}_j=\ldots=\mb{T}_J$, \ie\ do a recurrent
feature transformation.
However, this method has been observed slightly worse than the current
GFT in the performance.
Third, depending on the actual value of $J$, GFT might be slower in
computation than other gated networks which perform a single
transformation.
These three issues are left to our future work.

\section{Conclusions}
We have presented GFT, a simple but general neural language
grounding module for embodied agents.
GFT provides a unifying view of some existing language grounding
modules, and further generalizes on top of them.
The evaluation results on two challenging navigation environments
suggest that GFT can be easily adapted from one problem to another
robustly.
Although evaluated on navigation, we believe that GFT could
potentially serve as a general-purpose language grounding module for
embodied agents that need to follow language instructions in a variety
of scenarios.

\bibliography{corl2018.bbl}

\clearpage

\appendix

\section*{\textbf{Appendices}}

\section{Agent Architecture}
\label{app:ga2c}
The agent history $\mb{h}^{[t]}$ has two constituents: an action
history $\mb{h}_a^{[t]}$ summarizing previous taken actions and a
visual history $\mb{h}_m^{[t]}$ summarizing previous visual
experience.
We instantiate $f_{\theta}$ and $h_{\theta}$ as GRUs (Figure~\ref{fig:overview}):
\begin{equation}
  \label{eq:arch}
  \begin{array}{r@{\hskip 0.04in}l}
    \mb{h}_m^{[t]} &=
    GRU(m_{\theta}(\mb{o}^{[t]},\mb{l}), \mb{h}_m^{[t-1]}),\\
    \mb{h}_a^{[t]} &= GRU(a^{[t]}, \mb{h}_a^{[t-1]}),\\
    \mb{f}^{[t]} &= GRU(\mb{h}_m^{[t]}, \mb{h}_a^{[t-1]},
    \mb{f}^{[t-1]}),\\
    \mb{h}^{[t]} &= (\mb{h}_a^{[t]}, \mb{h}_m^{[t]}).\\
  \end{array}
\end{equation}

Our RL training design is synchronous advantage actor-critic
(ParallelA2C) \citep{Clemente2017}.
We run $N_{\text{agent}}$ agents in parallel with model parameters
shared among them to encourage exploration and reduce variance in the
policy gradient.
Each backpropagation is done with a minibatch of $N_{\text{batch}}$
time steps collected from all the agents, each agent contributing
$\frac{N_{\text{batch}}}{N_{\text{agent}}}$ time steps.
In either environment, every agent gets blocked until the network
parameters are updated with the current minibatch, after which it
forwards $\frac{N_{\text{batch}}}{N_{\text{agent}}}$ time steps again
with the updated model parameters.
To speed up training, we adopt an $n$-step temporal difference (TD)
when computing the advantage $A^{[t]}$ (Eq.~\ref{eq:advantage}), in a
forward manner similar to \citet{Mnih2016}.
Finally, we empirically set the discount $\gamma$ to $0.99$, the
entropy weight $\kappa$ to $0.05$, and the value regression weight
$\eta$ to $1.0$ throughout the experiments.

Sometimes an agent might provide fewer than
$\frac{N_{\text{batch}}}{N_{\text{agent}}}$ actions for each minibatch
due to the end of an episode, because once an agent hits an episode
end, it will be initialized in a new session for the next minibatch.
Thus the actual size of the minibatch might be smaller than
$N_{\text{batch}}$.
In the experiments, we set $N_{\text{agent}}=32$,
and $N_{\text{batch}}=128$.

The agent is trained purely from reward signals, without:
\begin{center}
\begin{compactenum}[1)]
  \item prior visual knowledge such as a pre-trained CNN,
  \item prior linguistic knowledge such as a parser, or
  \item any auxiliary task such as image reconstruction
    \citep{Das2018,Hermann17}, reward prediction \citep{Hermann17},
    or language prediction \citep{Hermann17,Yu2018}.
\end{compactenum}
\end{center}

\section{Method Details}
\label{app:model}
\tb{General.} The sentence embedding $\mb{l}_{\text{BoW}}$ is obtained
by sum-pooling word embeddings.
A word embedding has a length of $128$, except for the \tb{Concept}
method (see below).
The agent perceives $84\times 84$ egocentric RGB images
in \textsc{xworld3D} and $80 \times 80$ egocentric RGB images
in \textsc{xworld2D}.
Regardless of the image dimensions, the CNN has three convolutional
layers for processing the image: $(8, 4, 32)$, $(4, 2, 64)$, and $(3,
1, 64)$, where $(a,b,c)$ represents a layer configuration of $c$
filters of size $a\times a$ at a stride of $b$.
Each action is embedded as a vector of size $128$ before being input
to $GRU_{\theta}^a$ to generate the action history $\mb{h}_a$ which
also has $128$ units.
The other two recurrent layers $GRU_{\theta}^m$ and $GRU_{\theta}^f$
both have $512$ units, and both have an extra hidden layer of size
$512$ to preprocess their inputs.
The policy network is a two-layer MLP where the first layer has $512$
units and the second layer is a softmax for outputting actions.
The value network is a two-layer MLP where the first layer has $512$
units and the second layer outputs a scalar value without any
activation.
Unless otherwise stated, all the layer outputs are ReLU activated.

\tb{Concat.} Both $\mb{l}_{\text{BoW}}$ and $\mb{C}$ are projected to
a latent space of $512$ dimensions.

\tb{Gated.} The MLP for generating $\mb{l}_{\text{gate}}$ from $\mb{l}_{\text{BoW}}$
has two layers: the first layer has $128$ units and the second layer
has $D=64$ units which are sigmoid activated.

\tb{CGated.} The gate vector $\mb{l}_{\text{gate}}$ has $512$ units
which are sigmoid activated.

\tb{FiLM.} The MLP for generating $\lambda_d$ and $b_d$ has two
layers: the first layer has $128$ units and the second layer has
$D+1=65$ units without any activation.

\tb{Concept.} Because the sentence embedding $\mb{l}_{\text{BoW}}$ is
directly used as the $1\times 1$ filter, each word embedding has a length
of $D=64$.
The attention map and environment map are both ReLU activated.

\tb{EncDec.} This method has a slightly reorganized computational flow
compared to the one in Eq.~\ref{eq:arch}.
We refer the reader to the original paper \citep{Perez2018} for
details.
Except this, the configuration of each layer is the same with that of
its counterpart that can be found in \tb{Concat}.

\tb{GFT.} The MLP for generating the transformation matrix $\mb{T}_j$
from $\mb{l}_{\text{BoW}}$ has two layers: the first layer has $128$
units and the second layer has $D\times (D+1)=64\times 65=4160$ units.
The second layer has no activation.
For \tb{GFT-2}, we share the parameters of the first layers between
$\mb{T}_1$ and $\mb{T}_2$.
We set the activation function $g$ in Eq.~\ref{eq:staft} as ReLU.

\section{Vocabulary}
The following 8 spatial-relation words and 40 grammatical words are shared
between the \textsc{xworld2D} and \textsc{xworld3D}.

\begin{table}[h!]
    \centering
    \begin{tabular}{l|l}
        \tb{Spatial-relation} (8) & \tb{Grammatical} (40)\\
        \hline
        behind, & !, ., ?, and, anything, avoid,\\
        besides, & but, can, collect, could, destination,\\
        between, & do, done, end, except, go,\\
        by, & goal, grid, in, is, location,\\
        front, & move, navigate, not, object, of,\\
        left, & place, please, reach, target, that,\\
        near, & the, time, to, up, well,\\
        right. & will, wrong, you, your.\\
    \end{tabular}
\end{table}
The two environments have two different sets of object words:
\begin{table}[h!]
    \centering
    \begin{tabular}{@{}l|l@{}}
        \multicolumn{2}{@{}l}{\tb{Object}}\\
        \hline
        \textsc{xworld3D} (88) & \textsc{xworld2D} (115)\\
        \hline
        apples, backpack, barbecue, barrel, basket,
        &
        apple, armadillo, artichoke, avocado, banana, bat,
        \\
        basketball, bathtub, bed, bench, boiler,
        &
        bathtub, beans, bear, bed, bee, beet,
        \\
        books, bookshelf, bottle, bread, brush,
        &
        beetle, bird, blueberry, bookshelf, broccoli, bull,
        \\
        bucket, burger, cake, calender, camera,
        &
        butterfly, cabbage, cactus, camel, carpet, carrot,
        \\
        candle, car, carpet, cat, chair,
        &
        cat, centipede, chair, cherry, clock, coconut,
        \\
        chessboard, clock, comb, cooker, crib,
        &
        corn, cow, crab, crocodile, cucumber, deer,
        \\
        cup, dart, dog, dog-house, drawers,
        &
        desk, dinosaur, dog, donkey, dragon, dragonfly,
        \\
        drums, fan, fence, firehydrant, flashlight,
        &
        duck, eggplant, elephant, fan, fig, fireplace,
        \\
        flowers, fountain, gift, guitar, hair-dryer,
        &
        fish, fox, frog, garlic, giraffe, glove, horse,
        \\
        headphones, horse, iron, lamp, laptop,
        &
        goat, grape, greenonion, greenpepper, hedgehog,
        \\
        mailbox, milk, oven, pan, phone,
        &
        kangaroo, knife, koala, ladybug, lemon, light,
        \\
        photo, piano, pillow, plant, pool-table,
        &
        lion, lizard, microwave, mirror, monitor, monkey,
        \\
        puzzle, rabbit, rooster, scale, scissors,
        &
        monster, mushroom, octopus, onion, orange, ostrich,
        \\
        screen, slippers, sofa, speaker, squeezer,
        &
        owl, panda, peacock, penguin, pepper, pig,
        \\
        staircase, stove, sunglasses, table,
        &
        pineapple, plunger, potato, pumpkin, rabbit, racoon,
        \\
        table-tennis, toilet, towel, train,
        &
        rat, rhinoceros, rooster, seahorse, seashell, seaurchin,
        \\
        trampoline, trashcan, treadmill, tricycle,
        &
        shrimp, snail, snake, sofa, spider, squirrel,
        \\
        umbrella, vacuum, vase, wardrobe,
        &
        stairs, strawberry, tiger, toilet, tomato, turtle,
        \\
        wheelchair.
        &
        vacuum, wardrobe, washingmachine, watermelon,
        \\
        &
        whale, wheat, zebra.\\
    \end{tabular}
\end{table}

\begin{table}[t!]
\centering
\begin{tabular}{c|c|c|c}
    \tb{Level} & \tb{Map size ($X=Y$)} & \tb{number of goals per map}
    & \tb{number of obstacles per map}\\
    \hline
    1 & 3 & 2 & 0\\
    2 & 4 & 2 & 3\\
    3 & 5 & 2 & 6\\
    4 & 6 & 4 & 9\\
    5 & 7 & 4 & 12\\
    6 & 8 & 4 & 16\\
\end{tabular}
\caption{The curriculum used for training the agents.}
\label{tab:curriculum}
\end{table}

\section{Curriculum Learning}
To help the agent learn, we adopt curriculum
learning \citep{Bengio2009} to gradually increase the environment size
and complexity, according to the curriculum in
Table~\ref{tab:curriculum}.
The training always starts from level 1.
During training, the teacher maintains the average success rate of
each task type (Table~\ref{tab:tasks}), for a total of 200 most recent
sessions.
If at some point, all the five average success rates are above a
predefined threshold of $0.7$, then the teacher allows the agent to
enter the next level and resets the maintained rates.
The progress of this curriculum is computed separately for each of the
32 agents running in parallel.
The above curriculum applies to all the methods in both environments.
It should be noted that the training curves and test results in
Section~\ref{sec:exp} are computed for the final level with the
maximal difficulty, without being affected by the curriculum.

\begin{table}[t!]
  \centering
  \resizebox{\textwidth}{!}{
  \begin{tabular}{l|c|c|c|c|c|c|c|c|c|c}
    & \multicolumn{2}{c|}{\texttt{nav}}
    & \multicolumn{2}{c|}{\texttt{nav\_avoid}}
    & \multicolumn{2}{c|}{\texttt{nav\_bw}}
    & \multicolumn{2}{c|}{\texttt{nav\_dir}}
    & \multicolumn{2}{c}{\texttt{nav\_near}}
    \\
%    \cline{2-11}
    & 2D & 3D
    & 2D & 3D
    & 2D & 3D
    & 2D & 3D
    & 2D & 3D
    \\
    \hline
    \tb{Concat}
    & 46.1$\pm$4.9
    & 63.6$\pm$4.2
    & 94.1$\pm$1.3
    & 96.5$\pm$1.2
    & 72.2$\pm$6.2
    & 72.1$\pm$4.1
    & 46.2$\pm$26.1
    & 38.8$\pm$3.8
    & 47.1$\pm$11.7
    & 49.6$\pm$4.3
    \\
    \tb{Gated}
    & 49.5$\pm$4.1
    & 65.6$\pm$4.2
    & 94.4$\pm$1.8
    & 95.8$\pm$1.7
    & \tb{77.7}$\pm$4.8
    & 72.6$\pm$5.4
    & 66.4$\pm$5.0
    & 37.7$\pm$4.1
    & 55.5$\pm$4.7
    & 49.2$\pm$3.5
    \\
    \tb{CGated}
    & 47.1$\pm$3.2
    & 63.8$\pm$4.5
    & 94.6$\pm$1.6
    & 96.1$\pm$1.7
    & 75.5$\pm$4.1
    & 74.0$\pm$4.6
    & 62.8$\pm$4.5
    & 39.2$\pm$4.2
    & 56.2$\pm$4.0
    & 52.5$\pm$4.1
    \\
    \tb{FiLM}
    & 49.0$\pm$3.6
    & 68.2$\pm$3.1
    & 94.4$\pm$1.6
    & 97.4$\pm$1.1
    & 76.9$\pm$3.3
    & 74.0$\pm$3.8
    & 67.5$\pm$5.1
    & 43.3$\pm$3.1
    & 57.5$\pm$4.4
    & 56.6$\pm$2.8
    \\
    \tb{Concept}
    & 55.1$\pm$3.9
    & 55.6$\pm$17.1
    & 96.3$\pm$1.5
    & 95.0$\pm$4.6
    & 73.7$\pm$3.5
    & 53.3$\pm$21.0
    & 56.2$\pm$5.2
    & 27.6$\pm$12.7
    & 57.5$\pm$4.5
    & 43.7$\pm$14.7
    \\
    \tb{EncDec}
    & 40.3$\pm$7.3
    & 56.0$\pm$4.0
    & 89.4$\pm$6.6
    & 95.3$\pm$2.1
    & 66.9$\pm$10.3
    & 69.8$\pm$4.5
    & 20.6$\pm$25.4
    & 38.1$\pm$3.9
    & 43.3$\pm$12.6
    & 50.3$\pm$3.7
    \\
    \hline
    \tb{GFT-1}
    & 55.1$\pm$2.8
    & \tb{78.3}$\pm$3.0
    & 95.3$\pm$1.6
    & 97.8$\pm$1.0
    & 76.8$\pm$3.0
    & \tb{78.3}$\pm$4.4
    & 70.8$\pm$3.5
    & \tb{52.3}$\pm$5.4
    & 60.5$\pm$2.8
    & \tb{66.9}$\pm$3.8
    \\
    \tb{GFT-2}
    & \tb{60.9}$\pm$5.1
    & 76.2$\pm$3.6
    & \tb{96.8}$\pm$1.3
    & \tb{98.1}$\pm$0.9
    & 77.6$\pm$3.8
    & 76.3$\pm$4.1
    & \tb{71.4}$\pm$3.7
    & 51.8$\pm$5.8
    & \tb{65.1}$\pm$4.2
    & 66.8$\pm$4.3
    \\
  \end{tabular}
  }
  \caption{The evaluation results for 10k test sessions on $9\times 9$
    maps.
    The average navigation success rates are reported as percentages.
    The second number in each cell represents the standard deviation
    over twelve test runs, each run corresponding to one pass (out of
    three final passes) of one trained model (out of four random
    initializations).
  }
  \label{tab:generalization-9}
\end{table}

\begin{table}[t!]
  \centering
  \resizebox{\textwidth}{!}{
  \begin{tabular}{l|c|c|c|c|c|c|c|c|c|c}
    & \multicolumn{2}{c|}{\texttt{nav}}
    & \multicolumn{2}{c|}{\texttt{nav\_avoid}}
    & \multicolumn{2}{c|}{\texttt{nav\_bw}}
    & \multicolumn{2}{c|}{\texttt{nav\_dir}}
    & \multicolumn{2}{c}{\texttt{nav\_near}}
    \\
%    \cline{2-11}
    & 2D & 3D
    & 2D & 3D
    & 2D & 3D
    & 2D & 3D
    & 2D & 3D
    \\
    \hline
    \tb{Concat}
    & 44.2$\pm$3.2
    & 59.3$\pm$2.2
    & 94.9$\pm$1.6
    & 96.5$\pm$1.0
    & 70.3$\pm$6.7
    & 70.9$\pm$4.6
    & 43.8$\pm$24.9
    & 35.7$\pm$3.0
    & 47.6$\pm$12.1
    & 48.2$\pm$4.2
    \\
    \tb{Gated}
    & 45.9$\pm$4.6
    & 64.1$\pm$3.1
    & 94.4$\pm$1.6
    & 95.7$\pm$1.8
    & 76.7$\pm$3.9
    & 71.1$\pm$4.1
    & 63.5$\pm$4.0
    & 36.2$\pm$4.4
    & 56.2$\pm$5.6
    & 50.6$\pm$4.9
    \\
    \tb{CGated}
    & 44.9$\pm$3.8
    & 64.2$\pm$5.0
    & 93.9$\pm$1.5
    & 96.9$\pm$1.0
    & 76.3$\pm$4.9
    & 70.0$\pm$4.4
    & 63.4$\pm$4.2
    & 37.5$\pm$4.3
    & 56.6$\pm$3.8
    & 52.5$\pm$2.9
    \\
    \tb{FiLM}
    & 46.7$\pm$3.7
    & 66.1$\pm$4.2
    & 93.5$\pm$2.5
    & 97.6$\pm$1.2
    & 77.6$\pm$3.7
    & 72.4$\pm$3.7
    & 64.7$\pm$4.8
    & 41.2$\pm$4.8
    & 57.4$\pm$3.1
    & 56.4$\pm$2.7
    \\
    \tb{Concept}
    & 44.2$\pm$3.2
    & 55.5$\pm$18.3
    & 94.9$\pm$1.6
    & 94.5$\pm$5.9
    & 70.3$\pm$6.7
    & 50.9$\pm$21.2
    & 43.8$\pm$24.9
    & 24.1$\pm$11.0
    & 47.6$\pm$12.1
    & 40.2$\pm$15.6
    \\
    \tb{EncDec}
    & 37.8$\pm$6.0
    & 53.9$\pm$6.1
    & 91.4$\pm$5.5
    & 95.8$\pm$2.0
    & 66.7$\pm$12.3
    & 69.0$\pm$3.4
    & 18.6$\pm$23.7
    & 35.9$\pm$3.5
    & 41.9$\pm$11.0
    & 47.4$\pm$3.6
    \\
    \hline
    \tb{GFT-1}
    & 52.7$\pm$3.4
    & 78.1$\pm$4.3
    & 95.9$\pm$1.7
    & 98.3$\pm$0.9
    & \tb{78.7}$\pm$2.9
    & 75.5$\pm$4.3
    & 66.6$\pm$3.5
    & 51.6$\pm$3.8
    & 60.5$\pm$2.6
    & \tb{68.3}$\pm$3.5
    \\
    \tb{GFT-2}
    & \tb{60.7}$\pm$2.7
    & \tb{78.8}$\pm$3.2
    & \tb{97.3}$\pm$1.4
    & \tb{98.6}$\pm$1.3
    & 78.6$\pm$6.2
    & \tb{76.2}$\pm$3.5
    & \tb{69.2}$\pm$4.0
    & \tb{52.4}$\pm$4.5
    & \tb{65.7}$\pm$4.0
    & 68.2$\pm$5.2
    \\
  \end{tabular}
  }
  \caption{The evaluation results for 10k test sessions on $10\times
    10$ maps.
    The average navigation success rates are reported as percentages.
    The second number in each cell represents the standard deviation
    over twelve test runs, each run corresponding to one pass (out of
    three final passes) of one trained model (out of four random
    initializations).
  }
  \label{tab:generalization-10}
\end{table}

\begin{table}[t!]
  \centering
  \resizebox{\textwidth}{!}{
  \begin{tabular}{l|c|c|c|c|c|c|c|c|c|c}
    & \multicolumn{2}{c|}{\texttt{nav}}
    & \multicolumn{2}{c|}{\texttt{nav\_avoid}}
    & \multicolumn{2}{c|}{\texttt{nav\_bw}}
    & \multicolumn{2}{c|}{\texttt{nav\_dir}}
    & \multicolumn{2}{c}{\texttt{nav\_near}}
    \\
%    \cline{2-11}
    & 2D & 3D
    & 2D & 3D
    & 2D & 3D
    & 2D & 3D
    & 2D & 3D
    \\
    \hline
    \tb{Concat}
    & 36.2$\pm$5.2
    & 44.4$\pm$3.9
    & 96.3$\pm$1.5
    & 94.3$\pm$1.3
    & 64.9$\pm$7.0
    & 54.7$\pm$5.5
    & 42.0$\pm$23.0
    & 25.9$\pm$3.0
    & 41.2$\pm$12.8
    & 31.9$\pm$3.9
    \\
    \tb{Gated}
    & 40.3$\pm$5.2
    & 47.3$\pm$5.7
    & 96.0$\pm$1.8
    & 94.9$\pm$2.0
    & 70.5$\pm$4.6
    & 58.9$\pm$6.1
    & 58.4$\pm$5.0
    & 25.2$\pm$3.0
    & 48.7$\pm$3.6
    & 33.5$\pm$4.0
    \\
    \tb{CGated}
    & 35.9$\pm$3.0
    & 46.0$\pm$4.0
    & 95.3$\pm$2.0
    & 95.3$\pm$1.4
    & 72.4$\pm$5.1
    & 54.7$\pm$5.2
    & 54.0$\pm$4.0
    & 24.5$\pm$2.5
    & 46.7$\pm$3.9
    & 32.6$\pm$3.1
    \\
    \tb{FiLM}
    & 39.9$\pm$4.4
    & 53.2$\pm$4.6
    & 96.8$\pm$1.1
    & 95.5$\pm$1.0
    & 72.4$\pm$3.7
    & 59.1$\pm$5.4
    & 59.3$\pm$5.0
    & 28.3$\pm$3.4
    & 49.2$\pm$4.3
    & 36.6$\pm$3.9
    \\
    \tb{Concept}
    & 44.7$\pm$3.7
    & 42.9$\pm$15.7
    & 97.5$\pm$1.4
    & 95.1$\pm$3.0
    & 68.7$\pm$3.5
    & 36.1$\pm$16.4
    & 49.1$\pm$4.1
    & 17.4$\pm$9.6
    & 48.7$\pm$3.9
    & 27.6$\pm$11.2
    \\
    \tb{EncDec}
    & 31.7$\pm$5.9
    & 38.6$\pm$4.6
    & 93.5$\pm$4.3
    & 94.1$\pm$1.3
    & 62.3$\pm$11.6
    & 56.5$\pm$3.3
    & 16.1$\pm$21.7
    & 25.9$\pm$4.3
    & 38.0$\pm$11.5
    & 32.4$\pm$2.8
    \\
    \hline
    \tb{GFT-1}
    & 43.9$\pm$3.7
    & 62.6$\pm$4.0
    & 96.0$\pm$1.2
    & 95.9$\pm$1.5
    & \tb{74.3}$\pm$3.3
    & \tb{64.4}$\pm$4.6
    & 61.7$\pm$3.0
    & \tb{38.7}$\pm$4.0
    & 53.5$\pm$5.5
    & 49.5$\pm$3.6
    \\
    \tb{GFT-2}
    & \tb{51.5}$\pm$3.5
    & \tb{63.2}$\pm$5.2
    & \tb{98.0}$\pm$1.3
    & \tb{96.0}$\pm$1.5
    & 72.4$\pm$4.5
    & 64.2$\pm$5.4
    & \tb{64.1}$\pm$5.5
    & 37.5$\pm$3.5
    & \tb{58.9}$\pm$3.5
    & \tb{53.4}$\pm$4.2
    \\
  \end{tabular}
  }
  \caption{The evaluation results for 10k test sessions on $11\times
    11$ maps.
    The average navigation success rates are reported as percentages.
    The second number in each cell represents the standard deviation
    over twelve test runs, each run corresponding to one pass (out of
    three final passes) of one trained model (out of four random
    initializations).
  }
  \label{tab:generalization-11}
\end{table}

\section{Generalization to Larger Maps}
\label{app:generalization}
We evaluate the agent models, trained for $X=Y=8$ in
Section~\ref{sec:exp}, on three larger maps:
\begin{compactenum}[i)]
\item $X=Y=9$, with 6 goals and 20 obstacles,
\item $X=Y=10$, with 6 goals and 24 obstacles, and
\item $X=Y=11$, with 8 goals and 28 obstacles.
\end{compactenum}
The evaluation results are shown in
Table~\ref{tab:generalization-9}, Table~\ref{tab:generalization-10},
and Table~\ref{tab:generalization-11}, respectively.
Although the performance is not as good as on $8\times 8$ maps (except
for \texttt{nav\_avoid} whose chance performance tends to peak as
there are more goals on the map before the map size becoming too
large), the GFT agents achieve reasonable generalizations and still
greatly outperform the comparison methods.
This further demonstrates that our GFT agents are not trained to
memorize environments in specific settings.

\section{Navigation Examples}
\label{app:examples}
Below we show some navigation examples for the \tb{GFT-2} agents
trained in 2D and 3D.
For each session, we present four key frames on the navigation path.
More full-length navigation sessions are shown in a video demo
at \url{https://www.youtube.com/watch?v=bOBb1uhuJxg}.

\begin{figure}[t!]
  \begin{center}
  \resizebox{\textwidth}{!}{
    \begin{tabular}{@{}c@{}c@{}c@{}c@{}}
    \includegraphics[width=0.25\textwidth]{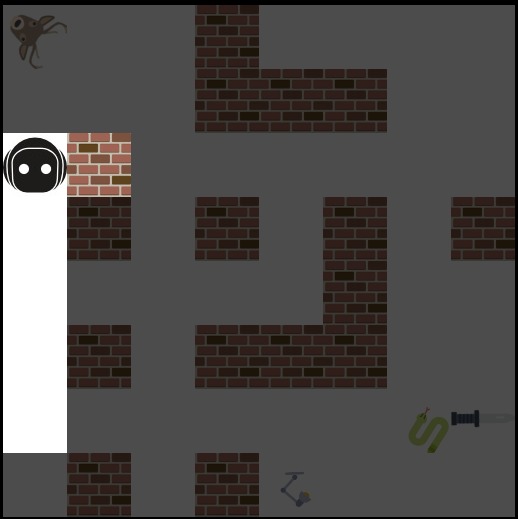}
    &
    \includegraphics[width=0.25\textwidth]{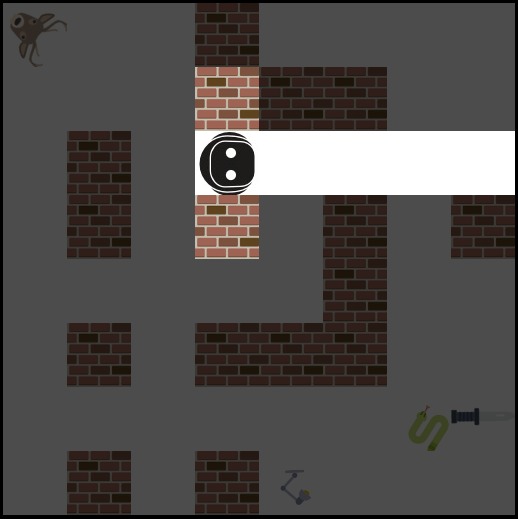}
    &
    \includegraphics[width=0.25\textwidth]{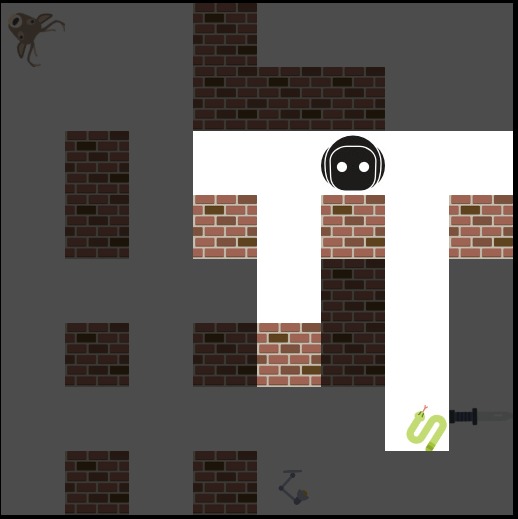}
    &
    \includegraphics[width=0.25\textwidth]{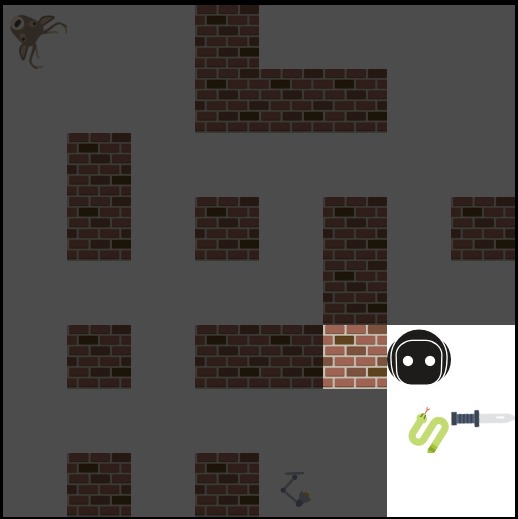}\\
    \multicolumn{4}{c}{``Anything except deer is the destination.''}\\
    \includegraphics[width=0.25\textwidth]{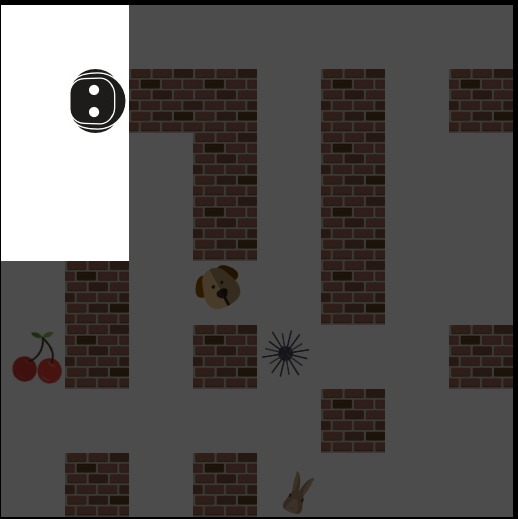}
    &
    \includegraphics[width=0.25\textwidth]{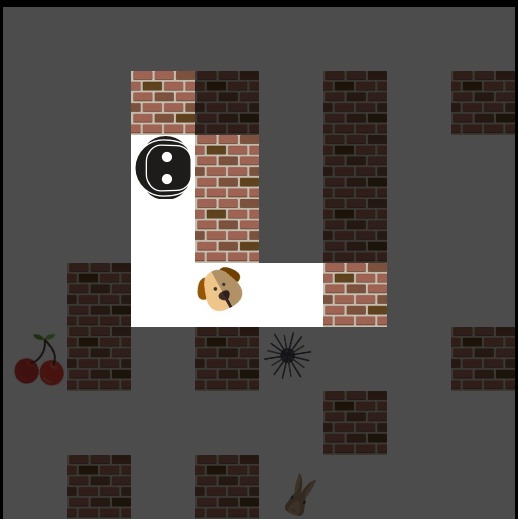}
    &
    \includegraphics[width=0.25\textwidth]{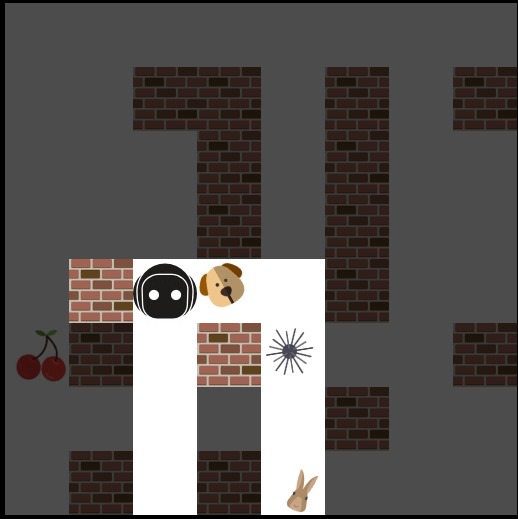}
    &
    \includegraphics[width=0.25\textwidth]{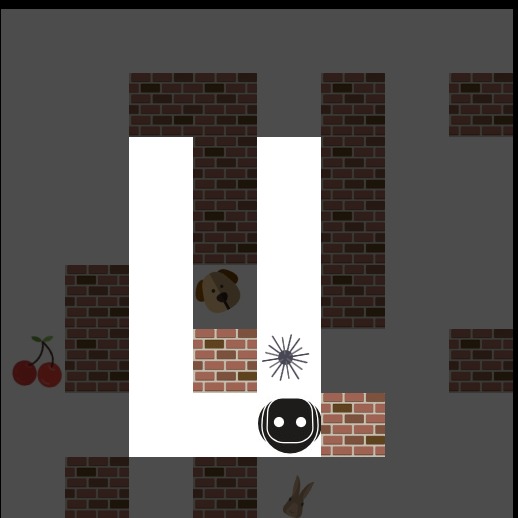}\\
    \multicolumn{4}{c}{``Will you go to the object by dog?''}\\
    \includegraphics[width=0.25\textwidth]{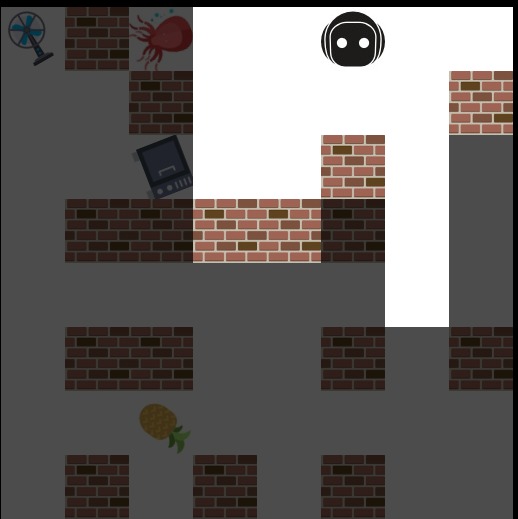}
    &
    \includegraphics[width=0.25\textwidth]{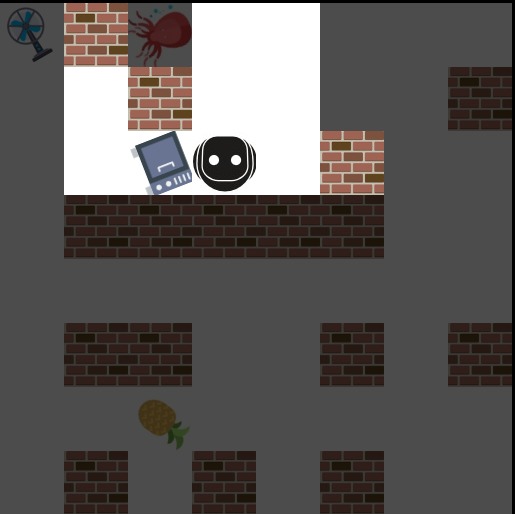}
    &
    \includegraphics[width=0.25\textwidth]{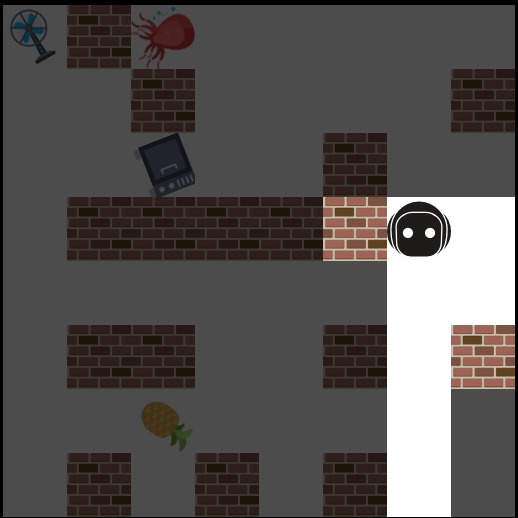}
    &
    \includegraphics[width=0.25\textwidth]{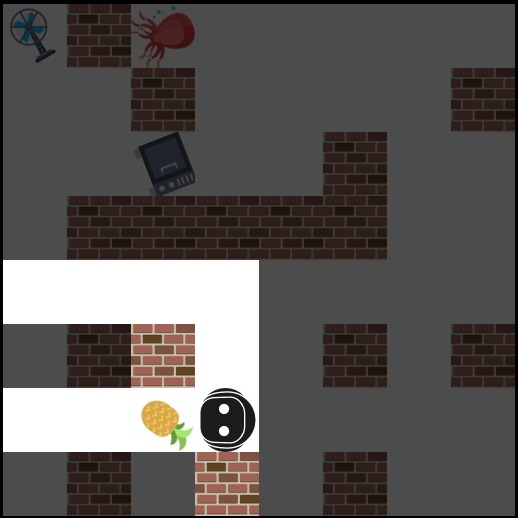}\\
    \multicolumn{4}{c}{``Pineapple is your destination.''}\\
    \includegraphics[width=0.25\textwidth]{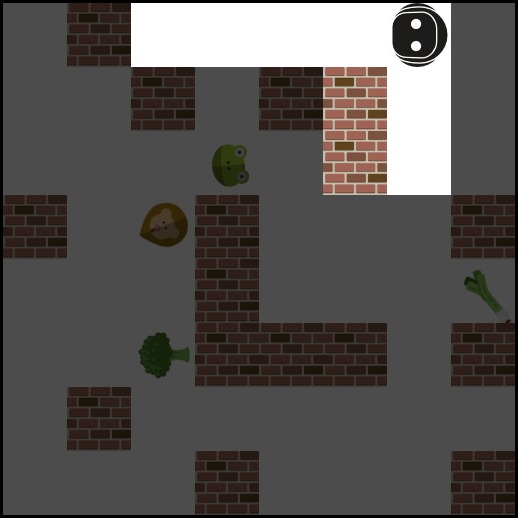}
    &
    \includegraphics[width=0.25\textwidth]{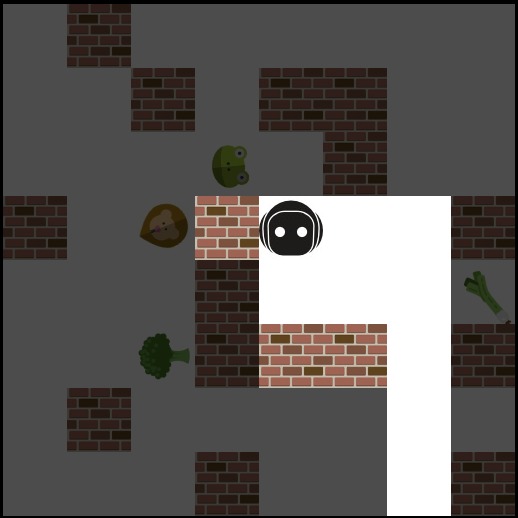}
    &
    \includegraphics[width=0.25\textwidth]{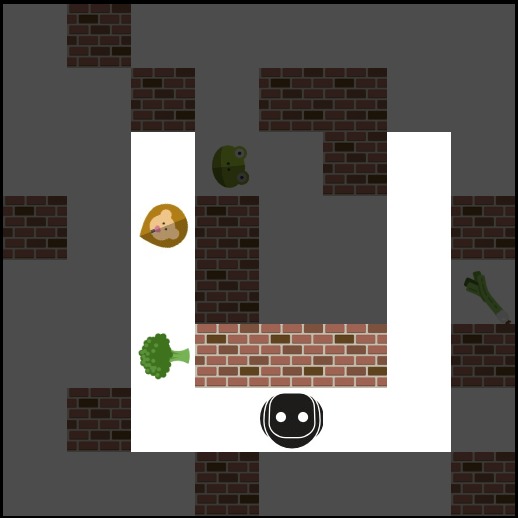}
    &
    \includegraphics[width=0.25\textwidth]{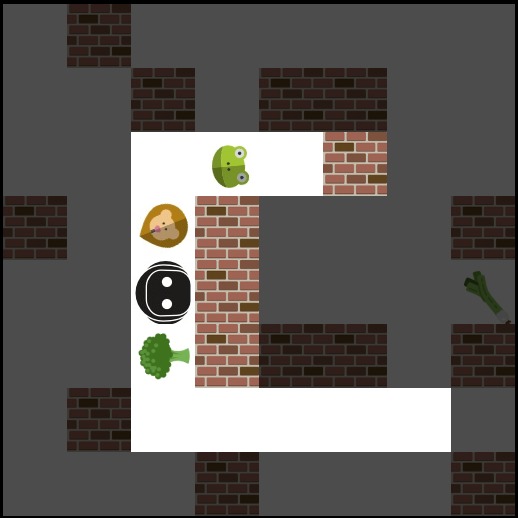}\\
    \multicolumn{4}{c}{``The location between lion and broccoli is
  your target.''}\\
    \includegraphics[width=0.25\textwidth]{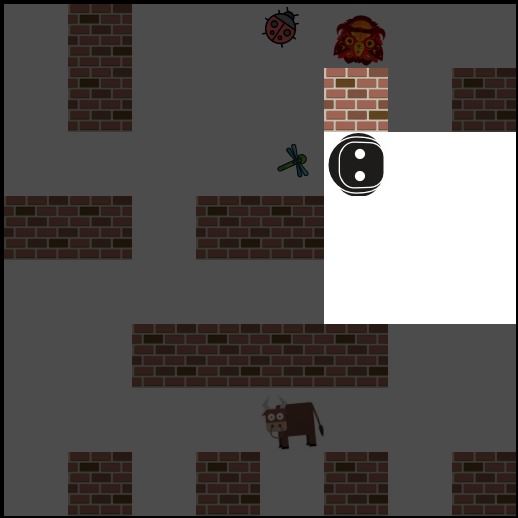}
    &
    \includegraphics[width=0.25\textwidth]{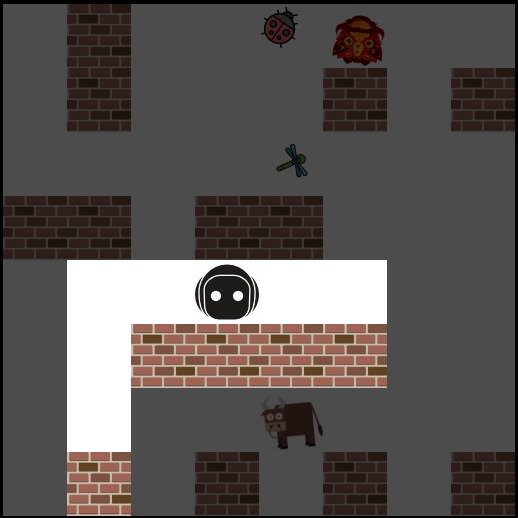}
    &
    \includegraphics[width=0.25\textwidth]{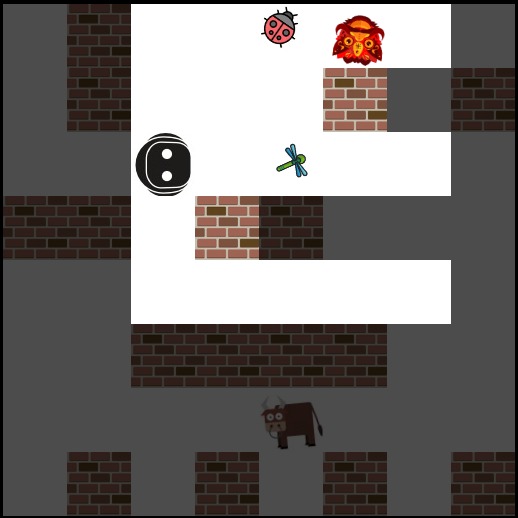}
    &
    \includegraphics[width=0.25\textwidth]{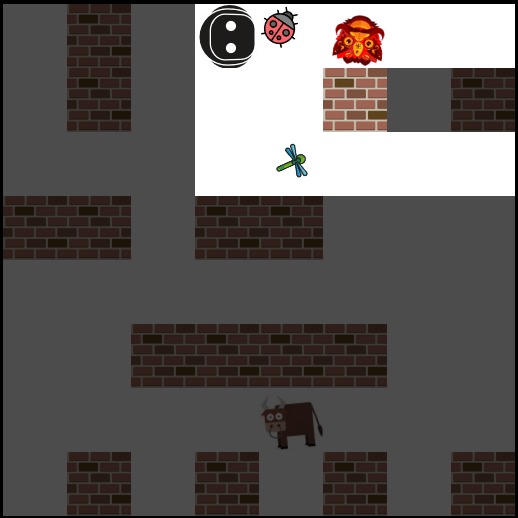}\\
    \multicolumn{4}{c}{``Collect the object in front of dragon.''}\\
    \end{tabular}
    }
  \caption{Five navigation examples for the \tb{GFT-2} agent
  trained in \textsc{xworld2D}.
  Four key frames in temporal order are shown in each example.
  During navigation, the agent is able to see only the highlighted
  regions.
  The shaded regions are for visualization purpose.
  }
  \label{fig:2d-examples}
  \end{center}
\end{figure}

\begin{figure}[t!]
  \begin{center}
  \resizebox{\textwidth}{!}{
    \begin{tabular}{@{}c@{}c@{}c@{}c@{}}
    \includegraphics[width=0.25\textwidth]{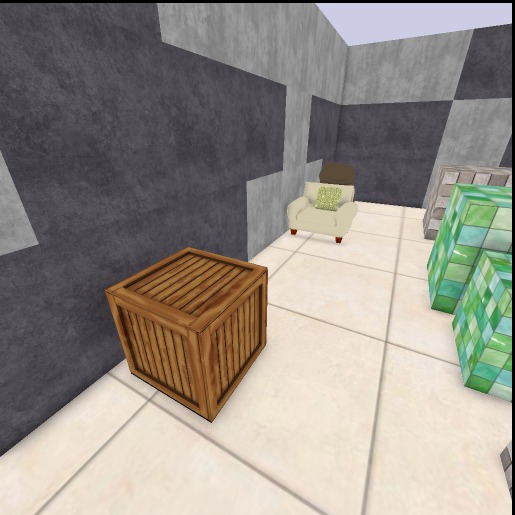}
    &
    \includegraphics[width=0.25\textwidth]{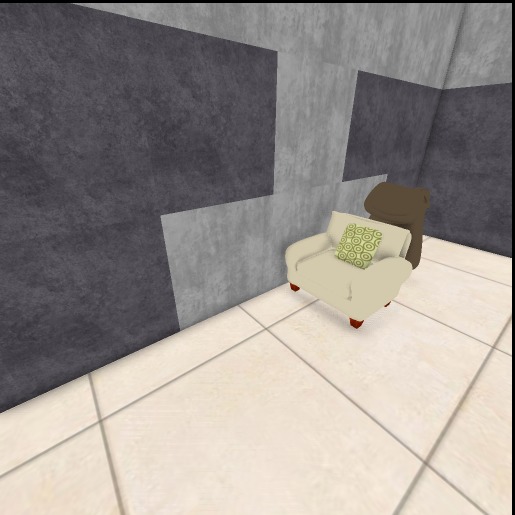}
    &
    \includegraphics[width=0.25\textwidth]{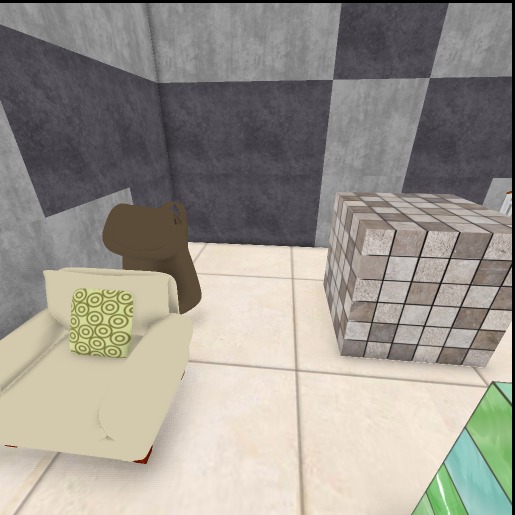}
    &
    \includegraphics[width=0.25\textwidth]{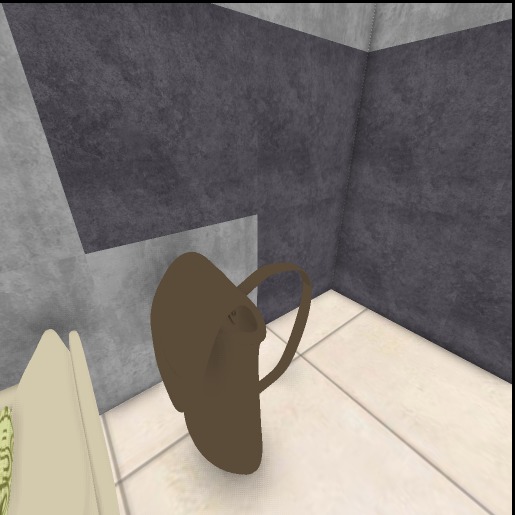}\\
    \multicolumn{4}{c}{``Reach the object that is to the right of sofa.''}\\
    \includegraphics[width=0.25\textwidth]{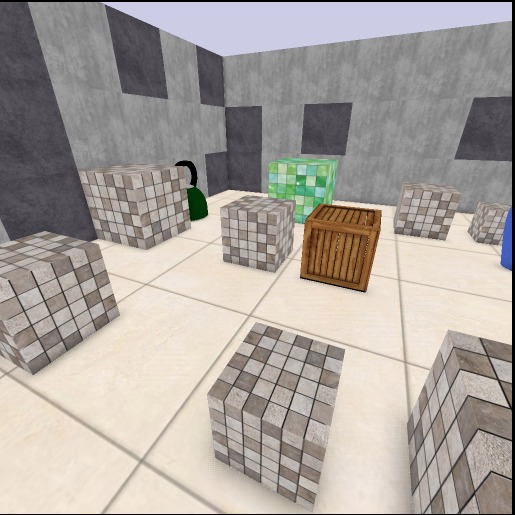}
    &
    \includegraphics[width=0.25\textwidth]{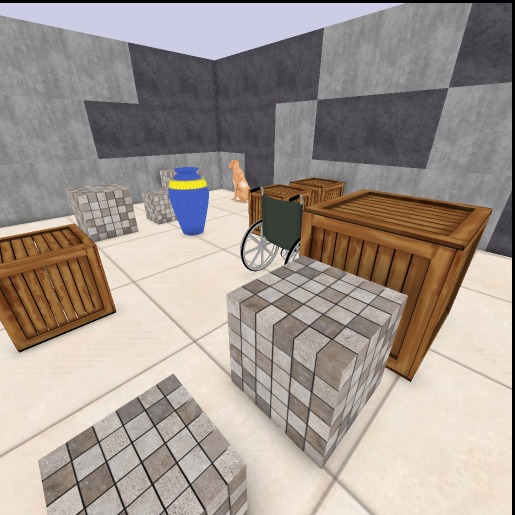}
    &
    \includegraphics[width=0.25\textwidth]{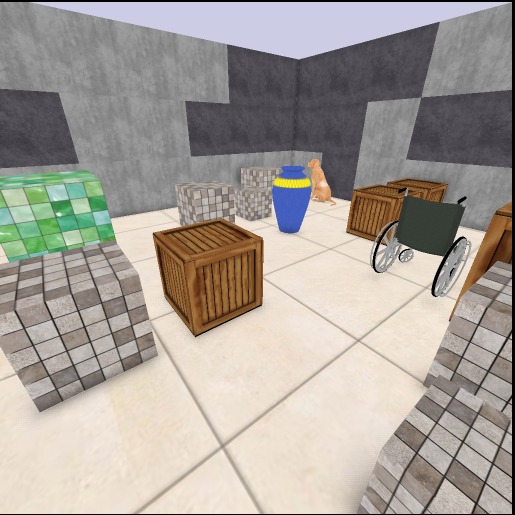}
    &
    \includegraphics[width=0.25\textwidth]{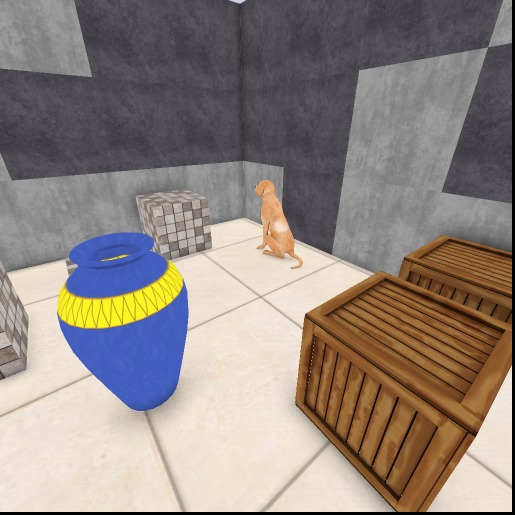}\\
    \multicolumn{4}{c}{``Move to the location between vase and
  wheelchair please.''}\\
    \includegraphics[width=0.25\textwidth]{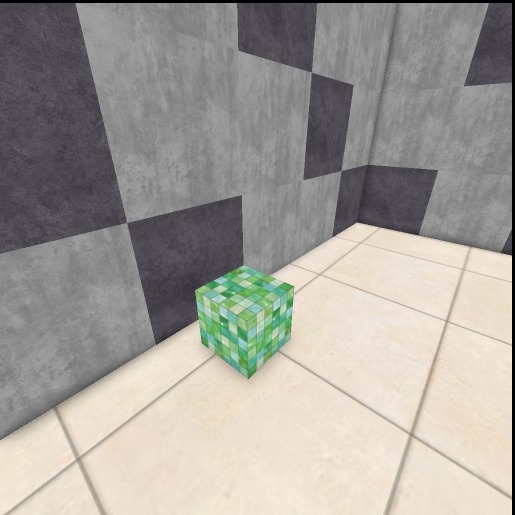}
    &
    \includegraphics[width=0.25\textwidth]{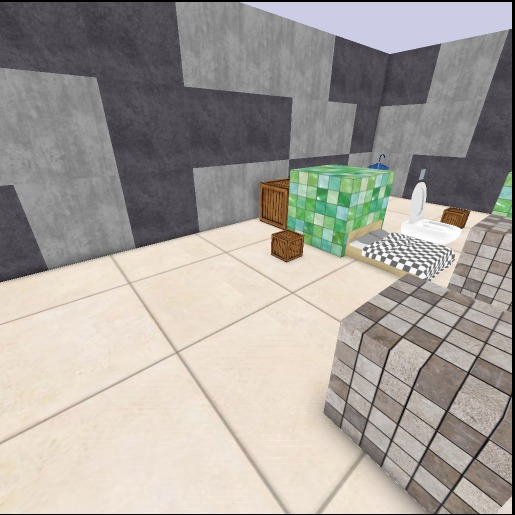}
    &
    \includegraphics[width=0.25\textwidth]{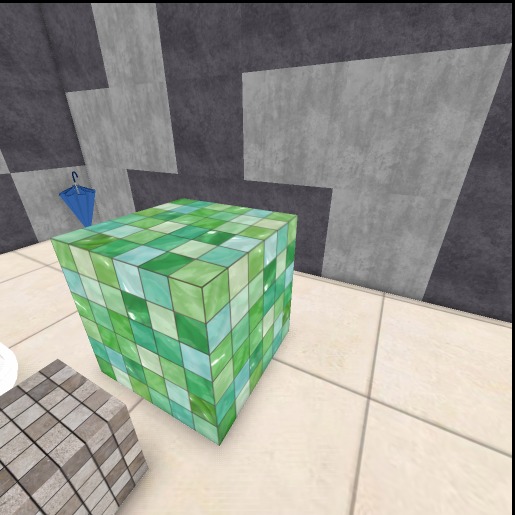}
    &
    \includegraphics[width=0.25\textwidth]{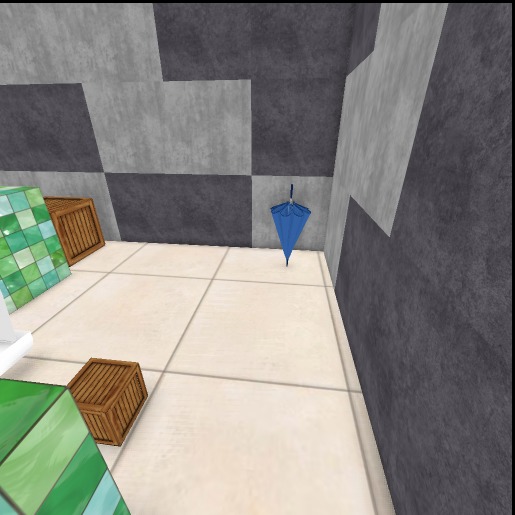}\\
    \multicolumn{4}{c}{``Could you please navigate to umbrella?''}\\
    \includegraphics[width=0.25\textwidth]{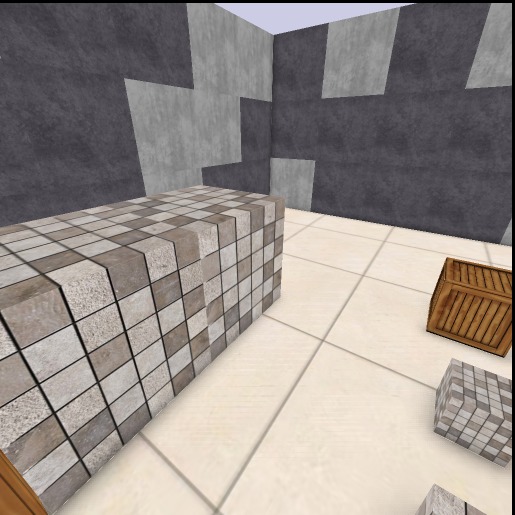}
    &
    \includegraphics[width=0.25\textwidth]{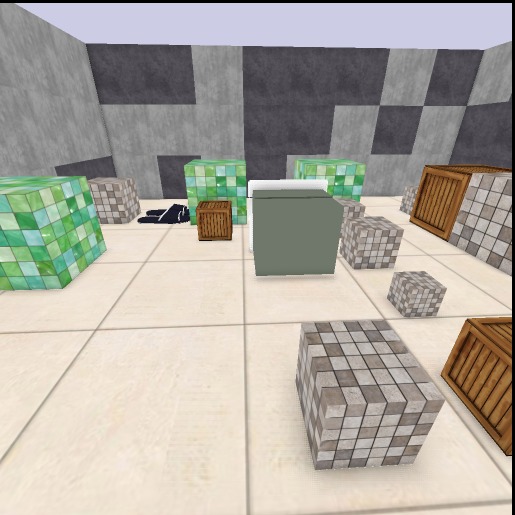}
    &
    \includegraphics[width=0.25\textwidth]{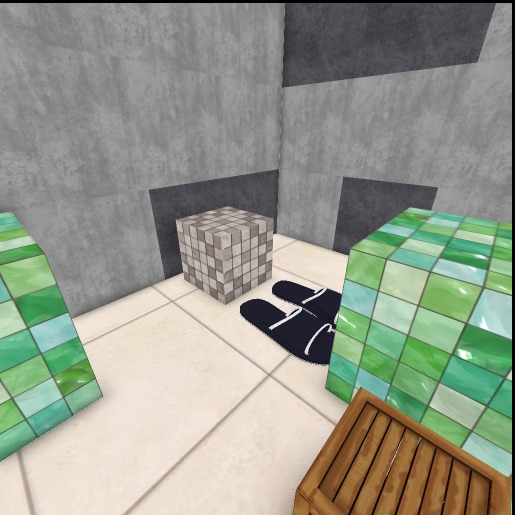}
    &
    \includegraphics[width=0.25\textwidth]{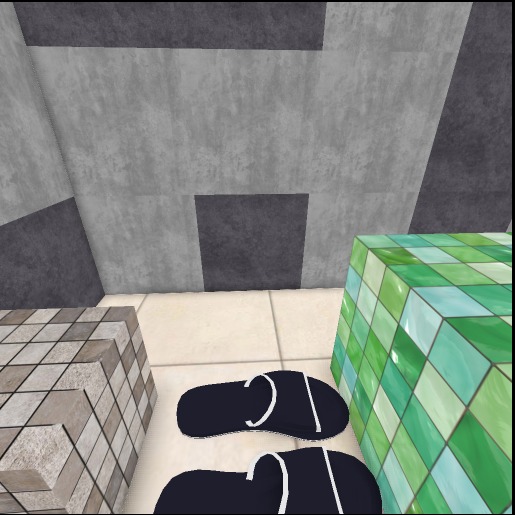}\\
    \multicolumn{4}{c}{``Please reach slippers.''}\\
    \includegraphics[width=0.25\textwidth]{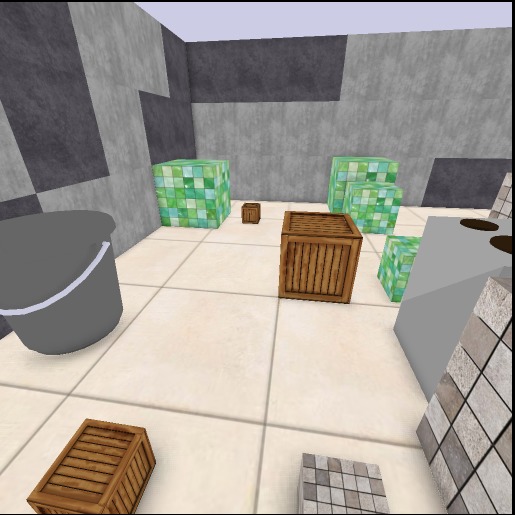}
    &
    \includegraphics[width=0.25\textwidth]{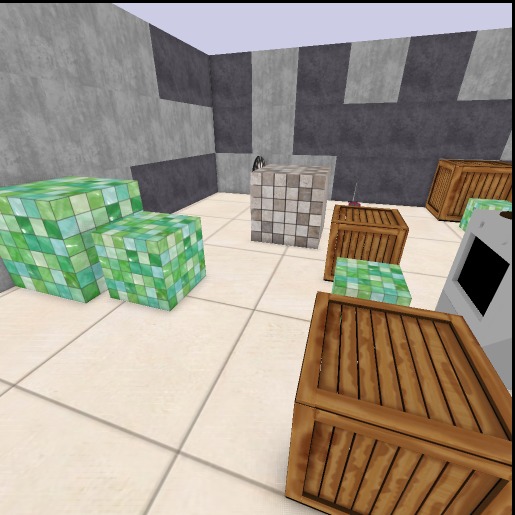}
    &
    \includegraphics[width=0.25\textwidth]{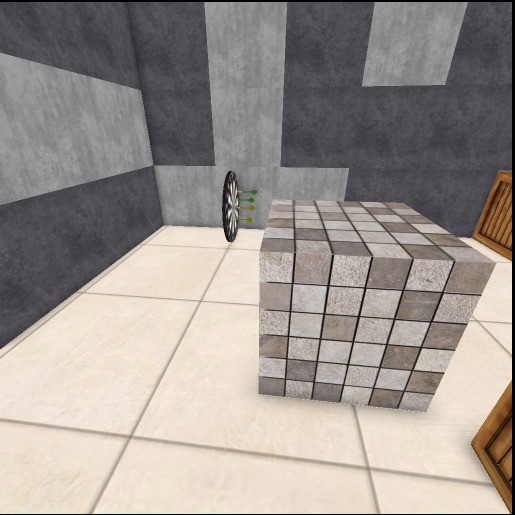}
    &
    \includegraphics[width=0.25\textwidth]{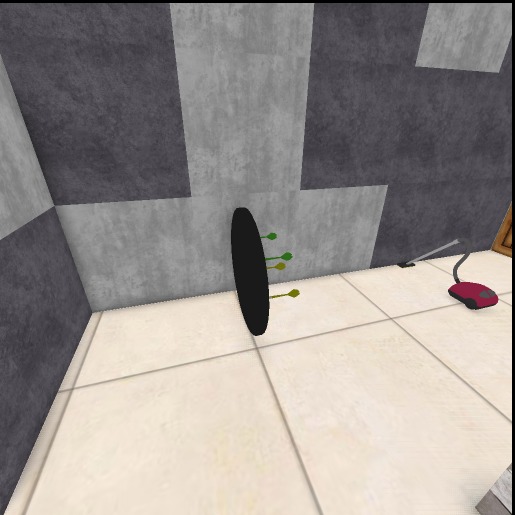}\\
    \multicolumn{4}{c}{``Go to the grid between dart and vaccum.''}\\
    \end{tabular}
    }
  \caption{Five navigation examples for the \tb{GFT-2} agent
  trained in \textsc{xworld3D}.
  Four key frames in temporal order are shown in each example.
  }
  \label{fig:3d-examples}
  \end{center}
\end{figure}

\clearpage
\section{Visualization of Transformation Matrices}
\label{app:visual}
\begin{figure}[!t]
  \begin{center}
  \resizebox{0.96\textwidth}{!}{
    \begin{tabular}{@{}c@{}c@{}}
      $\mb{T}_1$ \hspace{16ex} $\mb{T}_2$ & $\mb{T}_1$ \hspace{16ex} $\mb{T}_2$\\
      \includegraphics[width=0.48\textwidth]{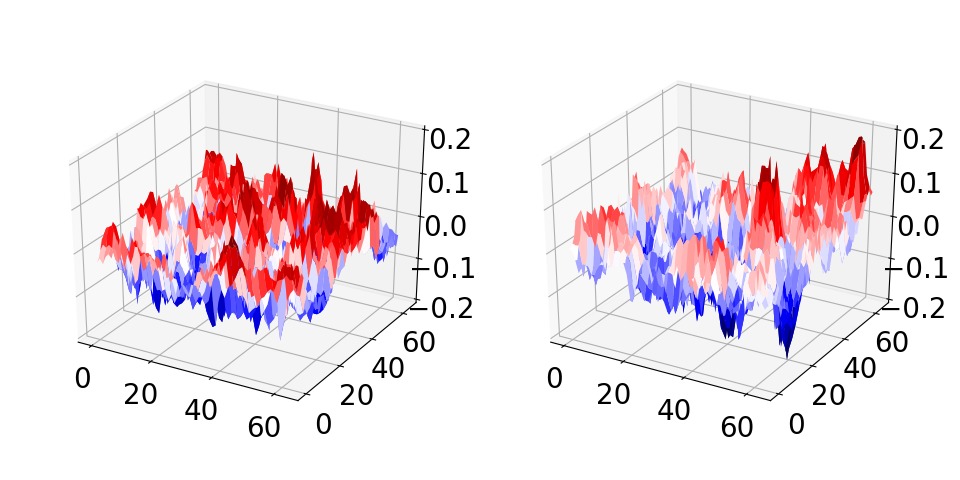}
      & \includegraphics[width=0.48\textwidth]{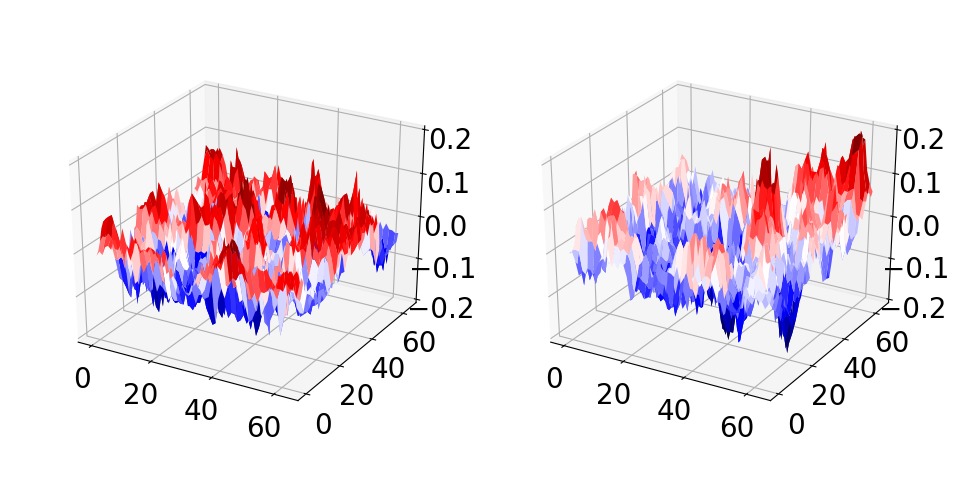}\\
      ``Will you avoid bottle?''
      & ``Anything but bottle is your destination.''\\
      \includegraphics[width=0.48\textwidth]{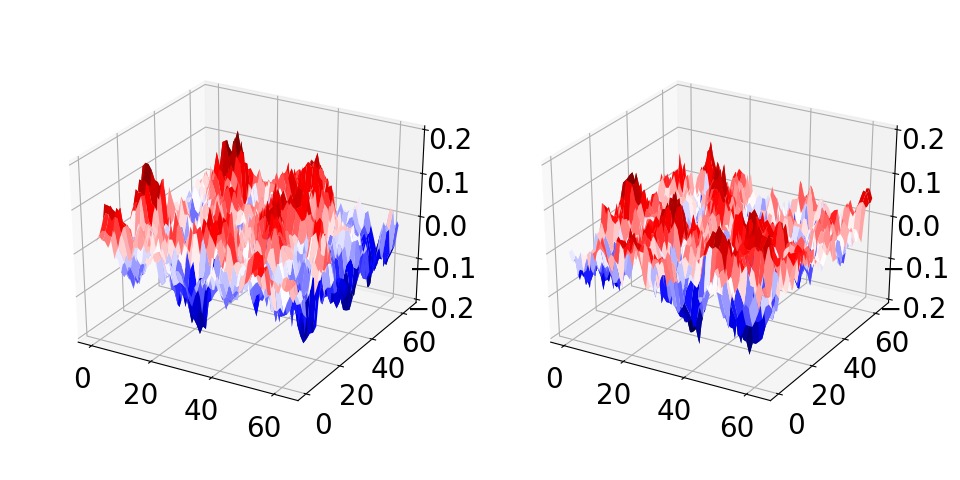}
      & \includegraphics[width=0.48\textwidth]{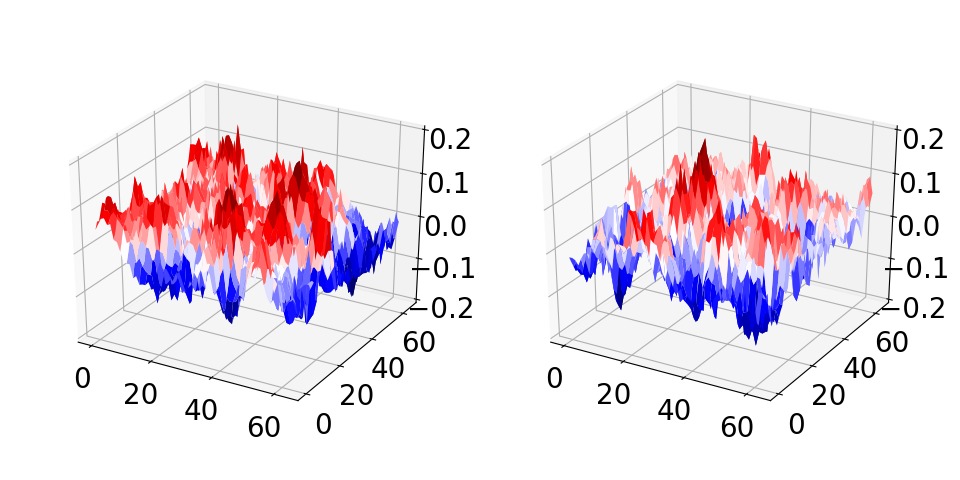}\\
      ``Please collect the object that is
      & ``Please collect the object\\
      in the front of vase.'' & to the left of vase.''\\
      \includegraphics[width=0.48\textwidth]{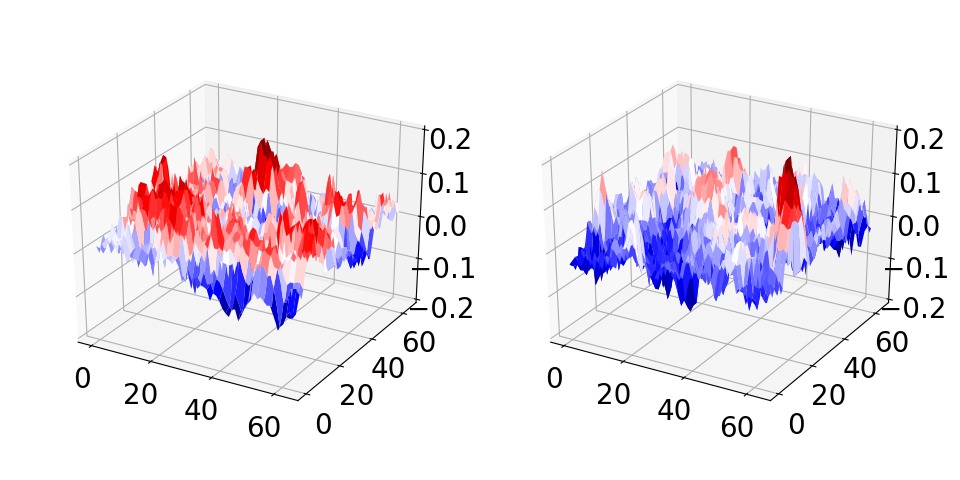}
      & \includegraphics[width=0.48\textwidth]{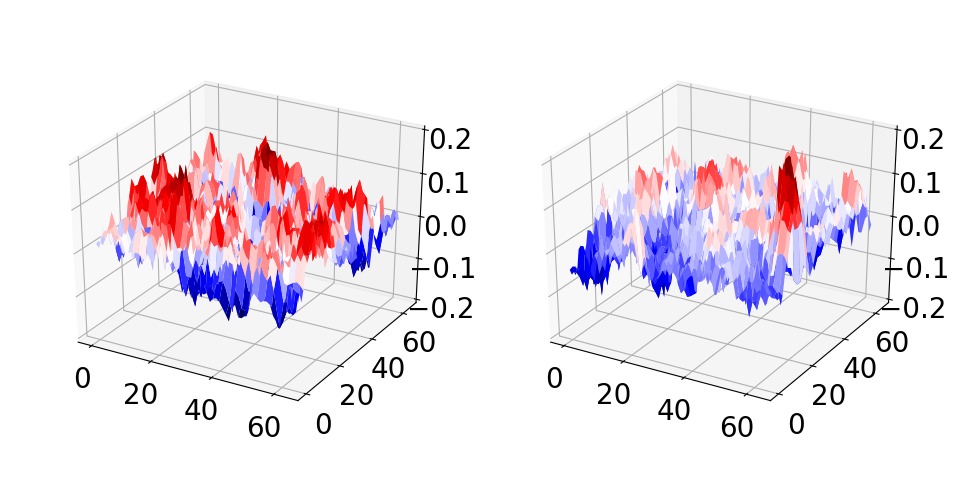}\\
      ``Go to the location between squeezer
      & ``Go to the location between vase\\
      and candle please.'' & and candle please.''\\
      \includegraphics[width=0.48\textwidth]{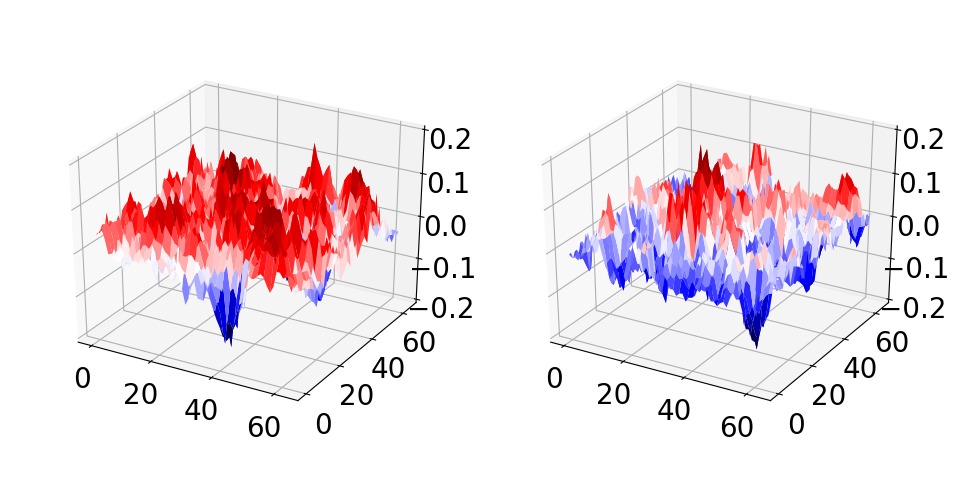}
      & \includegraphics[width=0.48\textwidth]{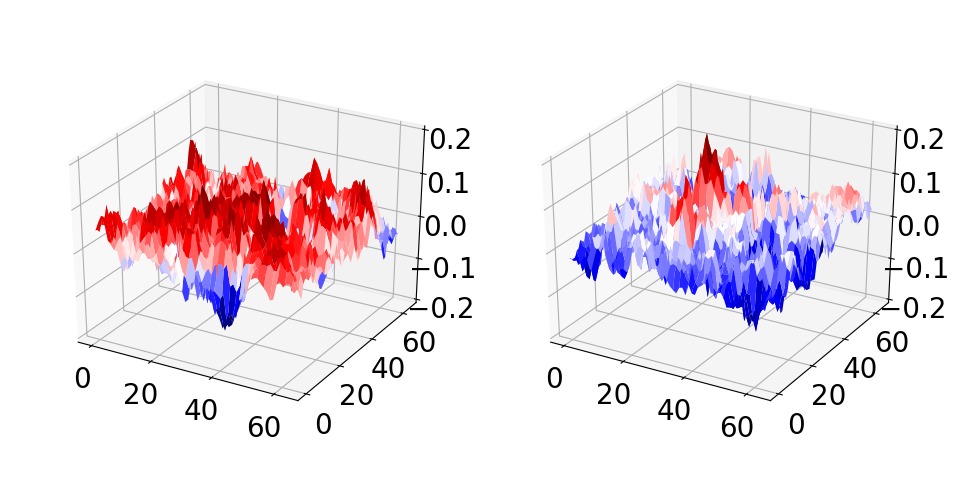}\\
      ``Go to the object left of bread.''
      & ``Go to the object left of basket.''\\
    \end{tabular}
    }
    \caption{Visualization of the transformation matrices computed by
  the trained \tb{GFT-2} agent for four example command pairs.
    For a better view, the matrices have been subtracted by the
    average $\mb{T}_1$ or $\mb{T}_2$ (computed from 5k randomly
    sampled commands) to remove the biases.
    Then each matrix is smoothed by a $7\times 7$ uniform kernel.
    }
    \label{fig:mats}
  \end{center}
\end{figure}

In this section, we visualize the transformation matrices of several
example commands received by the \tb{GFT-2} agent trained
in \textsc{xworld3D}.
We are particularly interested in comparing the $\mb{T}_1$ and
$\mb{T}_2$ of two different commands that have the same semantics or
have a minimal semantic difference.
Figure~\ref{fig:mats} shows the visualization results.
Unsurprisingly, we observe that
\begin{compactenum}[a)]
\item Two completely different commands with the same semantics (for
our problem) will yield almost identical transformation matrices
(Figure~\ref{fig:mats} row 1).
\item Two commands with a minimal difference will yield
matrices similar in general but differ in small places that capture
the difference (Figure~\ref{fig:mats} rows 2-4).
\end{compactenum}

\end{document}